\documentclass[pdflatex,iicol,sn-nature]{sn-jnl}

\usepackage{graphicx}%
\usepackage{multirow}%
\usepackage{amsmath,amssymb,amsfonts}%
\usepackage{amsthm}%
\usepackage{mathrsfs}%
\usepackage[title]{appendix}%
\usepackage{xcolor}%
\usepackage{textcomp}%
\usepackage{manyfoot}%
\usepackage{booktabs}%
\usepackage{algorithm}%
\usepackage{algorithmicx}%
\usepackage{algpseudocode}%
\usepackage{listings}%

\usepackage{url}            
\usepackage{booktabs}       
\usepackage{amsfonts}       
\usepackage{nicefrac}       
\usepackage{microtype}      
\usepackage{xcolor}         
\usepackage{makecell}
\usepackage{adjustbox}
\usepackage{pifont}
\usepackage{bbding}

\usepackage{tabulary,multirow,xspace}
\usepackage{fixmath,mathtools,nicefrac}
\usepackage{colortbl}
\usepackage{subcaption}
\usepackage{caption}
\usepackage{float}
\usepackage{array}
\usepackage{makecell}
\usepackage{ulem}
\usepackage{ragged2e}
\usepackage{bm}
\usepackage{amssymb}

\usepackage{lmodern}
\usepackage{fix-cm}
\usepackage{mathrsfs}
\usepackage{scalefnt}

\begin{document}

\title[Article Title]{Point-In-Context: Understanding Point Cloud via In-Context Learning}


\author[1]{\fnm{Mengyuan} \sur{Liu}}
\author*[2]{\fnm{Zhongbin} \sur{Fang}}\email{fanglaosi1107@gmail.com}
\author*[3]{\fnm{Xia} \sur{Li}}\email{ethlixia@gmail.com}
\author[3]{\fnm{Joachim M.} \sur{Buhmann}}
\author[4]{\fnm{Deheng} \sur{Ye}}
\author[5]{\fnm{Xiangtai} \sur{Li}}
\author[5]{\fnm{Chen Change} \sur{Loy}}

\affil[1]{\orgdiv{State Key Laboratory of General Artificial Intelligence}, \orgname{Peking University, Shenzhen Graduate School}, \orgaddress{\country{China}}}
\affil[2]{\orgdiv{School of Intelligent Systems Engineering}, \orgname{Sun Yat-sen University}, \orgaddress{\country{China}}}
\affil[3]{\orgname{Department of Computer Science}, \orgname{ETH Zurich},  \orgaddress{\country{Switzerland}}}
\affil[4]{\orgname{Tencent Inc.}, \orgaddress{\country{China}}}
\affil[5]{\orgname{S-Lab}, \orgname{Nanyang Technological University},  \orgaddress{\country{Singapore}}}

\abstract{
The rise of large-scale models has catalyzed in-context learning as a powerful approach for multitasking, particularly in natural language and image processing. However, its application to 3D point cloud tasks has been largely unexplored.
In this paper, we introduce Point-In-Context (PIC), a pioneering framework for 3D point cloud understanding that leverages in-context learning with a standard transformer architecture. PIC uniquely enables the execution of multiple tasks after a single, unified training phase, eliminating the need for fine-tuning.
To extend masked point modeling to 3D in-context learning, we introduce a Joint Sampling module, a simple yet effective technique that emphasizes the mapping relationship between input and target.
PIC treats both inputs and targets as coordinate-based, addressing the segmentation challenge by associating label points with 
pre-defined XYZ coordinates for each category. 
However, relying on such fixed label-coordinate assignments limits the model's ability to generalize to unseen domains.
To address this limitation, we further propose two innovative training strategies: In-Context Labeling and In-Context Enhancing. These strategies are integrated into an extended version of our model, named PIC++, which enhances dynamic in-context labeling and model training. 
Besides its multitask capability, PIC++ demonstrates superior performance and generalization across part segmentation datasets by employing dynamic in-context labels and regular in-context pairs.
Remarkably, PIC++, trained once without fine-tuning, can generalize effectively to unseen datasets and perform novel part segmentation through customized prompts.
Overall, PIC is a general framework that seamlessly integrates additional tasks or datasets through a unified data format via in-context learning.
Extensive experiments substantiate PIC's versatility and adaptability in handling diverse tasks and segmenting multiple datasets simultaneously.
The implementation of our framework is publicly available at \url{https://github.com/fanglaosi/Point-In-Context}.}

\keywords{In-context learning, part segmentation, multi-dataset, self-supervised learning
}

\maketitle

\begin{figure*}[ht]
\hsize=\textwidth
\centering
\includegraphics[width=\textwidth]{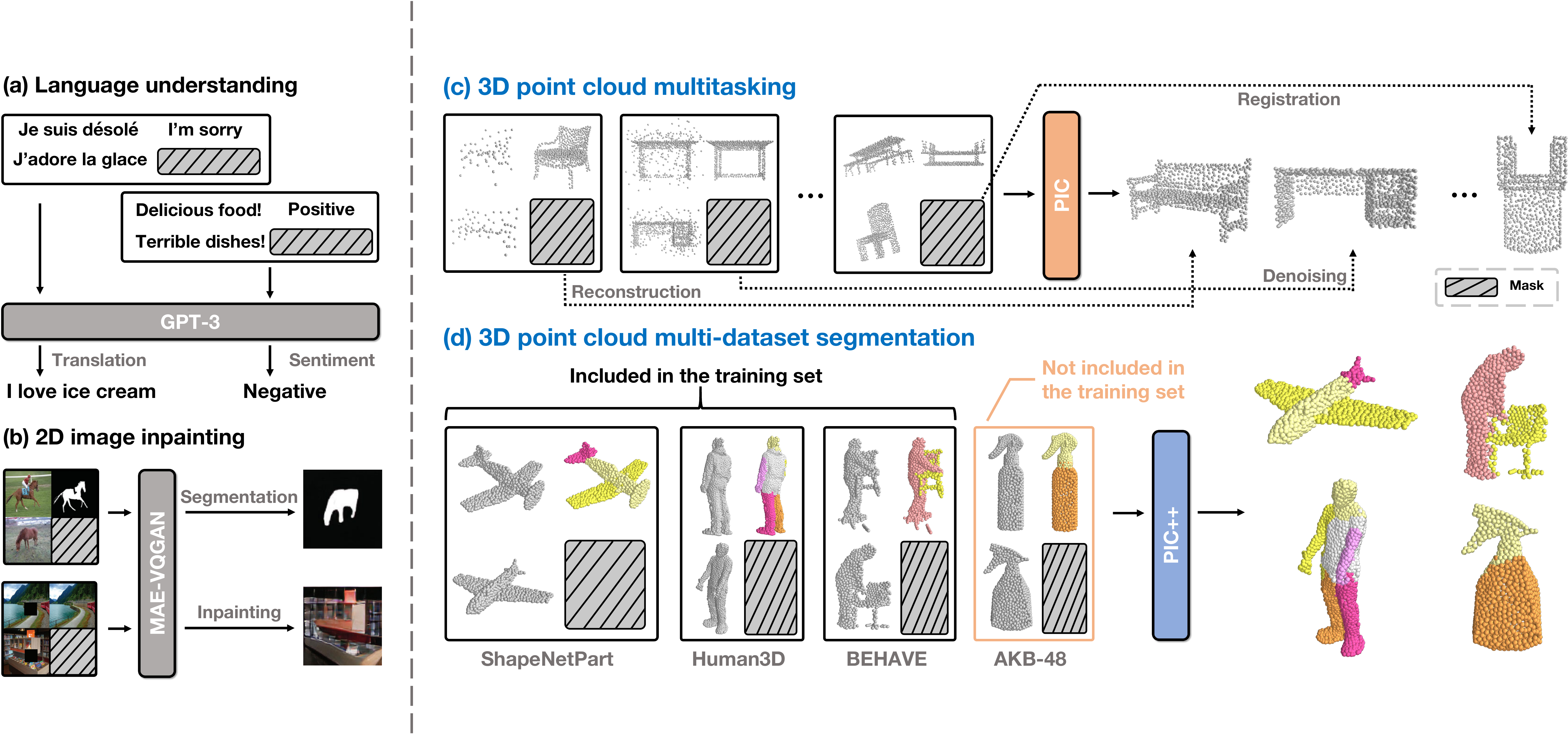}
\caption{
\textbf{Illustration of in-context learning for different tasks. }
\textbf{a)} In-context learning in NLP~\cite{gpt3}, with different text prompts for corresponding tasks: translation and sentiment analysis. 
\textbf{b)} In-context learning in 2D vision~\cite{visualprompt}, with 2D visual prompts for different tasks: segmentation and inpainting. 
\textbf{c)} Our proposed Point-In-Context~(PIC) for 3D point cloud multitasking, with 3D visual prompts for different tasks: reconstruction, denoising, registration, etc.
\textbf{d)} Our proposed Point-In-Context++~(PIC++) for multiple part segmentation datasets in 3D point clouds, including ShapeNetPart~\cite{shapenetpart}, Human3D~\cite{human3d}, BEHAVE~\cite{behave}, and AKB-48~\cite{akb48}. Note that our PIC++ can generalize to unseen segmentation dataset AKB-48, which is not included in the training set.
}
\label{teaser}
\end{figure*}

\section{Introduction}\label{sec:introduction}
In recent years, we have witnessed a remarkable surge in the development of large-scale models that boast an enormous number of parameters~\cite{flamingo,clip,scaling,opportunities,sam_iccv,OMGSeg,painter,wu2023open,learningtoprompt,learningtoretrieve,sic}.
These models have undergone pre-training on extensive datasets across the domains of Computer Vision (CV) and Natural Language Processing (NLP), showcasing their ability to tackle a multitude of tasks simultaneously and adapt to new challenges with ease. 
Pioneering models such as DALL-E for text-to-image generation~\cite{dell-e} and versatile language models like GPTs~\cite{gpt,gpt3} exemplify the cutting-edge advancements in this field.
Despite their remarkable capabilities, these sophisticated models face a significant barrier to adoption: the substantial computational resources required for their training process. Full fine-tuning~\cite{swintransformer,swintransformerv2,p-tuning} of these models is often prohibitive due to the high computational costs involved. Even more parameter-efficient tuning methodologies, such as prompt tuning~\cite{learningtoprompt,prefix,parameter}, can be impractical for a considerable portion of users who lack access to extensive computational resources. Fortunately, In-context learning~\cite{gpt3} offers a promising method to eliminate the need for fine-tuning down-stream tasks.

In-context learning, originating from NLP~\cite{gpt,learningtoretrieve,clip}, holds potential as the mainstream approach for efficient model adaptation and generalization. 
Unlike other methods that necessitate model parameter updates for new tasks, in-context learning incorporates domain-specific input-target pairs, known as in-context examples or prompts, into a query sample. 
In NLP, the prompt can be a machine translation pair or a sentiment analysis pair, as shown in Fig.~\ref{teaser}~(a). 
This allows the model to produce optimal results without requiring any parameter updates for previous tasks. 
Several works~\cite{visualprompt,painter} explore in-context learning in CV. The visual prompt study~\cite{visualprompt} is the first to adopt a trained neural network for filling missing patches in grid-like images, as shown in Fig.~\ref{teaser}~(b). 
Other studies~\cite{painter,seggpt} investigate the effect of visual prompts or generalization to more vision tasks.
The above methods adopt Mask Image Modeling~\cite{mae} (MIM) architecture for in-context task transfer. Inspired by MIM, Masked Point Modeling~\cite{pointmae} (MPM) has been proposed recently and is widely used in different point cloud tasks~\cite{i2pmae,pointm2ae}. 
To our knowledge, no work has explored in-context learning for 3D point cloud understanding using the MPM framework. 
In this study, we propose \textbf{Point-In-Context~(PIC)}, the first in-context learning framework for 3D point cloud understanding, realizing one training, one model, and processing multiple tasks, which is shown in Fig.~\ref{teaser}~(c). 
Additionally, we propose the extended version of PIC, named \textbf{Point-In-Context++~(PIC++)}, a unified in-context model specifically designed for multi-entity part segmentation tasks, which is shown in Fig.~\ref{teaser}~(d).

To systematically evaluate the 3D in-context learning capability, which serves as the foundation for both PIC and PIC++, we first
propose a multitask benchmark leveraging commonly available datasets such as ShapeNet~\cite{shapenet} and ShapeNetPart~\cite{shapenetpart}, including four common point cloud tasks: point cloud reconstruction, denoising, registration, and part segmentation. 
Meanwhile, we benchmark several representative baselines, including task-specific models customized for each task and multitask models equipped with a shared/pre-trained backbone with multiple task heads. Then, to tackle the multi-task benchmark mentioned above, we present the base framework fundamental in-context learning for 3D point clouds.

A straightforward extension of the 3D MPM framework to point clouds for in-context learning may
encounter two primary obstacles.
Firstly, the conventional 3D MPM approach of employing position embeddings may introduce information leakage.
As shown in Fig.~\ref{fig:technical_contri}~(a), previous pre-training pipelines embed positional information with the center point coordinates of all patches, even for those that are masked out~(invisible). 
Since the patches masked out in the target should be invisible during inference, such an operation will cause information leakage, which does not satisfy the requirements in the in-context learning setting. 
Secondly, unlike the structured nature of 1D word embeddings or 2D images, 3D point cloud data, which are inherently unordered, present the risk of having their positional information disarrayed when partitioned into a patch sequence.
In response to these issues, we propose a simple yet effective solution, termed the Joint Sampling~(JS) module. This technique involves recording the indices of sampled center points and employing the K-nearest neighbor strategy to simultaneously sample both the input and target. 
In addition, we explore two distinct baselines for PIC, which encompass separating inputs and targets akin to the Painter~\cite{painter} strategy and concatenating inputs and targets analogous to the MAE~\cite{mae} approach for reconstruction.
It is worth noting that, unlike the conventional focus of few-shot learning on a single specific task, our objective with PIC is to explore the in-context learning capabilities of models. 
This empowers the model to execute tasks aligned with the 3D prompt without updating model parameters, broadening its applicability and versatility within 3D point cloud understanding.
We also adopt the Objaverse dataset~\cite{objaverse} (800K) to verify that integrating large-scale data into the training set can enhance performance on in-domain or out-of-domain datasets.

For the part segmentation task, PIC converts each category into a point containing XYZ coordinates and forms a pre-defined label point map for segmentation tasks. However, this approach faces three obstacles in segmentation. 
1) Trained to fit fixed label contracts, the essence of in-context learning, thus part segmentation is isolated from other tasks.
2) When multiple datasets are involved, a large set of labels needs to crowd into a limited coordinate space, making the label regression more challenging.
3) Fixed label-coordinate assignment sacrifices the possibility of accommodating novel labels when facing unseen datasets.

To address large-scale and diverse part segmentation tasks within a single training, we further propose an extended version of PIC, PIC++, which includes the In-Context Labeling and the In-Context Enhancing training strategies. 
\textit{In-Context Labeling} is the key for PIC++ to generalize to other unseen datasets.
It revolutionizes the approach to labeling by replacing fixed label coordinates with dynamic context-aware label points. 
By randomly assigning coordinates to part categories within each point cloud and aligning prompts with shared part semantics, PIC++ fosters a deeper understanding of semantic information while enhancing scalability and extensibility across various segmentation datasets.
Furthermore, the \textit{In-Context Enhancing} strategy provides more diverse point cloud pairs in various corruption operations, enabling the model to learn more robust mapping relationships from the prompts.
Emphasizing in-context semantic information within prompts, PIC++ learns to generalize effectively across diverse datasets.
During inference, PIC++ dynamically adapts the label-point assignments for each query point cloud based on a temporary label bank from prompts, ensuring robust performance even on unseen datasets and providing convenience for seamlessly introducing other datasets into our dataset.

To explore stronger performance and better generalization of PIC on segmentation tasks, we create a new benchmark related to part segmentation in point clouds. 
The new benchmark includes ShapeNetPart~\cite{shapenetpart}, Human3D~\cite{human3d}, BEHAVE~\cite{behave}, and AKB-48~\cite{akb48}. 
We process these datasets into a format suitable for part segmentation and integrate them into large-scale Multi-Entity In-Context Datasets, consisting of 140K samples.

Our previous work is published in NeurIPS-2023~\cite{pic}(spotlight), and we make more significant contributions in this extension to make the first solid exploration of 3D point cloud in-context learning: 

\begin{itemize}
    \item Within the 3D in-context learning framework, we further propose PIC++. 
    In particular, we propose In-Context Labeling and In-Context Enhancing to improve performance and generalization capability in 3D point cloud part segmentation tasks.
    Furthermore, our PIC++ can seamlessly integrate additional segmented datasets without redundant label points.
    \item We establish the Multi-Entity Segmentation benchmark, which comprises four available point cloud datasets on human and object segmentation, including ShapeNetPart, Human3D, BEHAVE, and AKB-48. 
    Our goal is to fully evaluate the performance of models trained jointly on multiple segmentation datasets/tasks, as well as their generalization to unseen datasets/tasks.
    \item We conduct extensive experiments on the multi-tasking and multi-dataset segmentation benchmark to validate our PIC++. Compared to other models, PIC++ achieves SoTA performance. Furthermore, we show that PIC++ can produce excellent results on out-of-domain part segmentation datasets or customized part segmentation tasks, which makes our work more applicable in real-world scenarios.
\end{itemize}

\section{Related Work}

\noindent
\textbf{3D Point Cloud Analysis.} With the increased large-scale 3D datasets~\cite{modelnet40,rs,modelnet-o,wu2023omniobject3d}, research on 3D point cloud analysis mainly explores deep neural network architectures~\cite{guo2020deep}. 
Several pioneer works adopt the point-based methods, with Multi-Layer Perception~(MLP), to directly process each point into high-dimensional embeddings, including PointNet~\cite{pointnet} and PointNet++~\cite{pointnet++}. 
Then, PointNeXt~\cite{pointnext} improves PointNet++ via well-designed training strategies and an inverted residual bottleneck design. 
Meanwhile, graph-based methods~\cite{dgcnn,lpdnet,hierarchicalgraph,lin2021learninggraph} leverage geometric properties. They propose different dynamic kernels with dynamic graph convolution architectures.
In particular, DGCNN~\cite{dgcnn} designs EdgeConv to compute each layer's output dynamically and achieves comparable results with point-based methods.
Later, with the rise of vision transformers~\cite{vit,transformer}, some works~\cite{pointtransformer, pct, pointr} adopt transformer-based approaches to build the global context between each point. 
Point Transformer series~\cite{pointtransformer,pointtransformerv2,pointtransformerv3} explore modified transformer layers from different aspects, including a new vectorized self-attention mechanism, an efficient and accurate architecture, and extensive traverse functions. 
As a concurrent work, PCT~\cite{pct} presents a global-local self-attention layer.
In addition, several works integrate the vision-language models~\cite{pointclip,clip2} and 2D image~\cite{p2p,openscene,vg4d} features via transformer architectures for 3D point cloud analysis.
Several works~\cite{lai2022stratified,Schult23ICRA} explore transformer architecture for scene-level point cloud understanding with larger receptive fields.
Recently, there have also been several works exploring state space models~\cite{mamba,han2024mamba3d,pcm,pointmamba1}. 
In particular, MAMBA3D~\cite{han2024mamba3d} and PCMamba~\cite{pcm} show great potential for 3D point cloud analysis and achieve comparable results with transformer-based models but with global receptive fields with linear complexity. 
However, these approaches are only designed for a single task for point cloud representation learning, so they cannot be used directly for in-context learning.
Our work builds the first step in exploring in-context learning in the 3D point cloud multitasking and 
delves into the generalization of in-context learning in multi-dataset segmentation.

\noindent
\textbf{Masked X Modeling.} By masking partial data and predicting the missing contents, Masked X Modeling acts as an effective pre-training method in both language modeling and computer vision tasks.
In particular, starting from BERT~\cite{bert} and GPTs~\cite{gpt,gpt3,language}, Masked Language Modeling~(MLM) methods have greatly enhanced natural language processing performance by pre-training and fine-tuning diagrams. 
In computer vision, BEiT~\cite{beit} is the first to propose matching image patches with discrete tokens via d-VAE~\cite{dvae} and pre-train a standard vision transformer~(ViT~\cite{vit}) using Masked Image Modeling~(MIM). 
Later, MAE~\cite{mae} shows that we can reconstruct the raw pixel values of masked tokens and achieve high efficiency with a high mask ratio.
For 3D point clouds, several works~\cite{pointbert,pointmae,pointrae,mae3d,acl,lee2022ap} design a Masked Point Modeling~(MPM) pre-training framework. 
For example, Point-BERT~\cite{pointbert} adopts a BERT-like architecture, while PointMAE~\cite{pointmae} transfers the MAE-like framework for pre-training via reconstructing point cloud patches.
Meanwhile, several works develop improved MPM-style frameworks via specific technologies, such as hierarchical sampling~\cite{pointm2ae}, cross-modal distillation~\cite{i2pmae,pimae,recon,act}, etc. 
These methods all use standard transformer networks to process 3D point clouds and achieve competitive performance on various downstream tasks. 
Our approach follows a similar MPM pipeline but explores the in-context ability of point transformers and MPM. 
To our knowledge, this has not been investigated previously. 
In particular, via in-context learning, we can develop a unified model via a 3D task prompt, further widening the research scope in the 3D point cloud.

\begin{table*}[t]
\caption{\textbf{Comparison of ShapeNet In-Context Datasets and Multi-Entity In-Context Datasets with other common datasets.} The underline represents the datasets we use.}
\setlength\tabcolsep{1mm}
\centering
\scalebox{0.6}{
\begin{tabular}{l|c|ccc}
\toprule
Dataset                                          & Tasks                                                                                    & In-context pair & Objects & Categories \\ \hline
\underline{ShapeNet}~\cite{shapenet}                                         & 3D point cloud classification                                                            & -               & 51k     & 55         \\
ModelNet~\cite{modelnet40}                                         & 3D point cloud classification                                                            & -               & 12k     & 40         \\
3D-Future~\cite{3dfuture}                                        & 3D shape recognition, pose estimation, reconstruction, retrieval, Texture synthesis      & -               & 16k     & 34         \\
ABO~\cite{abo}                                              & Single-view 3D reconstruction, Material prediction, Object retrieval                     & -               & 8k      & 63         \\
ScanObjectNN~\cite{scanobjectnn}                                     & 3D point cloud classification                                                            & -               & 15k     & 15         \\
\underline{ShapeNetPart}~\cite{shapenetpart}                                     & 3D point cloud part segmentation                                                         & -               & 17k     & 50         \\
\underline{Human3D}~\cite{human3d}                                          & 3D instance segmentation, 3D semantic segmentation, Multi-human body part segmentation   & -               & 57k     & 6          \\
\underline{BEHAVE}~\cite{behave}                                           & Humen-object interaction tracking, 3D reconstruction, Pose and shape estimation          & -               & 15k     & 20         \\
\underline{AKB-48}~\cite{akb48}                                           & Object pose estimation, Object reconstruction, Object manipulation                       & -               & 2k      & 48         \\ \hline
\textbf{ShapeNet In-Context Datasets}                     & 3D reconstruction, denoising, registration, part segmentation                            & \Checkmark               & 217K    & 55         \\
\textbf{Multi-Entity In-Context Datasets} & Object part segmentation, Human part segmentation, Human-object interaction segmentation & \Checkmark               & 140K    & 94         \\ \bottomrule
\end{tabular}}
\vspace{-1em}
\label{tab:comparison_of_datasets}
\end{table*}

\noindent
\textbf{Multi-Task Learning.} Multi-task learning~(MTL)~\cite{multitasksurvey} aims to learn multiple tasks jointly so that the knowledge contained in a task can be leveraged by other tasks and increase the average performance.
Common methods design novel training strategies~\cite{hipro,mpa} or model structures~\cite{crossstich,learningtobranch,gnas,progressive,branched,controllable} to share information and achieve mutual benefits across multiple target tasks. 
In 3D, several works~\cite{unsupervisedmultitask,venvision3d} rely on heavily-designed multiple task heads~\cite{robust,gpanet} and the combination of various loss functions~\cite{jldnet,dnact}. 
For example, Hassani et al~\cite{unsupervisedmultitask} propose an unsupervised multitask model for jointly learning point and shape features in point clouds. 
To extract highly effective features, GPA-Net~\cite{gpanet}, comprising an encoder and multitask decoders, proposes a novel graph convolution kernel termed GPAConv.
The aforementioned methods achieve multitask learning by incorporating multiple task heads. 
In contrast, our approach explores a new way to achieve multitask learning by task prompting. 
We integrate several tasks into a unified input-target space, enabling training across various tasks within the same framework without task-specific head designs.

\noindent
\textbf{In-Context Learning.} This concept is a novel learning paradigm within large language models like GPT-3~\cite{gpt3}, facilitating inference on novel tasks by conditioning inputs on specific input-output pairs, known as "prompts". 
This paradigm empowers users to customize a model's output to align with their downstream datasets without modifying the often inaccessible internal model parameters~\cite{learningtoprompt,learningtoretrieve}. 
Recent advancements~\cite{flamingo,explanation} in language models underscore the effectiveness of in-context learning across diverse language tasks such as machine translation and sentiment analysis. 
Moreover, several studies~\cite{exploring,incontextscene,personalize,sam_iccv,goodexample,instructmemore,selfprompting,lvm} delve into extending in-context learning to computer vision tasks.
Notably, Visual Prompt~\cite{visualprompt} stands out as the pioneering pure 2D vision model, pre-trained to reconstruct masked patches within images comprised of academic figures and infographics. 
Subsequently, Painter~\cite{painter} extends this concept by generalizing visual in-context learning to image painting across different prompts. 
Later, SegGPT~\cite{seggpt} expands upon Painter by enabling generalized one-shot segmentation.
Some works~\cite{icldiff,generative,instructcv,wang2024explore} introduce in-context learning in diffusion-based generative models~\cite{diff}.
In contrast, our approach investigates the influence of 3D prompts on in-context learning within point clouds, proposing novel baselines for benchmarking 3D in-context learning efficacy.
In addition, we further enhance the 3D prompt design on dense prediction tasks like segmentation.

\section{3D Point Cloud In-Context Datasets}
\label{sec:in-context_datasets}

In this section, we first model the in-context learning paradigm in  3D point clouds as the 2D in-context learning methods do~(Sec.~\ref{sub_sec:modeling}). Then we introduce a novel multitasking benchmark for 3D in-context inference based on ShapeNet~\cite{shapenet} and ShapeNetPart~\cite{shapenetpart} datasets~(Sec.~\ref{sub_sec:shapenet_dataset}). Lastly, to further explore the in-context capability for 3D part segmentation, we establish a new multi-dataset joint training benchmark, including ShapeNetPart~\cite{shapenetpart}, Human3D~\cite{human3d}, BEHAVE~\cite{behave}, and AKB-48~\cite{akb48}~(Sec .~\ref{sub_sec:seg_dataset}). Tab.~\ref{tab:comparison_of_datasets} shows the comparison of our two novel benchmarks with other commonly used datasets.

\subsection{Modeling In-Context Learning in 3D Point Cloud}
\label{sub_sec:modeling}

\noindent
\textbf{In-Context Learning in 2D.} Previous works, such as Visual Prompt~\cite{visualprompt} and Painter~\cite{painter} combine two pairs of images that perform the same task into a grid-like image and randomly mask portions, following MAE~\cite{mae}.
During training, these models~\cite{visualprompt,painter} take two pairs of images $I_i$ and $I_j$ as inputs and form two pairs, including a reference pair (also termed prompt), $R^{k}_{i} =\{I_i, T^k_i\}$ and a query pair, $Q^k_j = \{I_j, T^k_j\}$. 
$R^k_i$ is the task prompt containing one image $I_i$ and its corresponding target $T^k_i$. 
$Q^k_j$ represent the current input image $I_j$ and its target $T^k_j$, which perform the same task with the prompt pair. 
Here, $k$ represents the task index. 
When $i=j$, we term the prompt as an \textit{ideal prompt}. 
During inference, only task example pairs and a query image are provided. 
They are combined into a grid-like image with a quarter mask to mask $T^k_j$, and a pre-trained model is used to restore the missing parts, as shown in Fig.~\ref{teaser}~(b).

\noindent
\textbf{Modeling In-Context Learning in 3D.} Following the same spirit of 2D in-context learning, for the first time, we design a similar procedure for 3D in-context learning. 
During training, each input sample contains two pairs of point clouds, including prompt point clouds $R^{k}_{i} =\{P_i, T^k_i\}$ and query point clouds $Q^k_j = \{P_j, T^k_j\}$, which perform the same task as in 2D in-context learning. 
Each pair consists of an input point cloud and its corresponding target point cloud for the given task. 
Similar to PointMAE~\cite{pointmae}, we adopt the farthest point sampling~(FPS) and K-nearest neighbor (KNN) techniques to convert the point clouds into a sentence-like data format. 
These point patches are subsequently encoded into tokens.
During inference, the input point cloud is a combination of prompt and query point cloud, while the target point cloud consists of an example target along with masked tokens, as shown in Fig.~\ref{teaser}(c) and Fig.~\ref{teaser}(d). 
Consequently, we obtain predictions corresponding to the position of the mask. 
Given an input, the models can output corresponding targets following various prompts.

\subsection{ShapeNet In-Context Datasets}
\label{sub_sec:shapenet_dataset}
To establish the first benchmark for 3D in-context learning, we carefully curate datasets and define task specifications, as there is no prior benchmark. 
Initially, we collect samples from representative datasets like ShapeNet~\cite{shapenet} and ShapeNetPart~\cite{shapenetpart}, transforming them into the ``input-target'' format as stated in Sec.~\ref{sub_sec:modeling}. 
To augment the sample size for part segmentation, we apply random operations such as \textit{point cloud perturbation, rotation, and scaling} to ShapeNetPart. 
This yields an extensive dataset with 217,454 samples across four tasks. Each sample includes an input point cloud and its corresponding task-specific target. 
We standardize the XYZ coordinates of inputs and outputs to the range between $[-1, 1]$, and unify all samples as pure point clouds in $(N,3)$ shape. 

\noindent
\textbf{Reconstruction.} 
The task is to reconstruct a dense point cloud from a sparse one. During processing, points are discarded by setting them to zero rather than reducing their number, to maintain consistent input lengths between input and target. To evaluate the reconstruction ability, we introduce the corruption level $R_{rec}$, where only $N \times R_{rec}$ points are input.

\noindent
\textbf{Denoising.} In this task, the input is a point cloud mixed with Gaussian noise $X \sim N(0,1)$. The objective is to remove the noise and produce a clear and distinct object shape. We also introduce the corruption level $R_{den}$ for input point clouds. This indicates that the input point cloud contains $N \times R_{den}$ noisy points. 

\noindent
\textbf{Registration.} This task aims to restore a rotated point cloud to its initial orientation. We assume that during both training and inference, the query point cloud and the prompt are synchronized in terms of the rotation angle. 
To prevent interference with the denoising and reconstruction tasks, we configure the registration task's output to include both an upright and an upside-down point cloud. 
We also introduce five corruption levels for the evaluation on this task.

\noindent
\textbf{Part Segmentation.} The objective of this task is to segment an object into multiple components, typically ranging from 2 to 6. Traditionally, each point is assigned a $C$-dimensional one-hot code for classification. However, for in-context learning, maintaining the input and output within the same space, which only contains XYZ coordinates, is crucial. Hence, we transform part segmentation from a classification task to a regression task.
To accomplish this, we generate a part label map consisting of $P$ discrete points representing part labels, derived from converting the $P$ labels into $P$ points with XYZ coordinates, where $P$ represents the total number of part categories. Upon obtaining the model's predicted result, we determine the part to which each output point belongs by calculating its distance to the part label map. This classifies the point as belonging to the part label point closest to it.

\subsection{Multi-Entity In-Context Datasets}
\label{sub_sec:seg_dataset}

To further improve the performance and generalization of PIC to segmentation tasks, we collect more datasets related to part segmentation in point clouds. 
These datasets include ShapeNetPart~\cite{shapenetpart}, originally used for synthetic object part segmentation, Human3D~\cite{human3d} for human body-part segmentation, BEHAVE~\cite{behave} for human-object interaction segmentation, and AKB-48~\cite{akb48} for real-collected object part segmentation. 
These datasets contain point clouds widely used in 3D human-machine interaction scenarios~\cite{nam2023cyclic}, including human, object, and human-object interaction point clouds.
We process these datasets into a format suitable for part segmentation and integrate them into large-scale Multi-Entity In-Context Datasets, consisting of approximately 140k samples. 
Moreover, to test the model's generalization, we exclude AKB-48 from the training set and use it as a one-shot generalization test dataset.

\noindent
\textbf{ShapeNetPart}~\cite{shapenetpart} dataset consists of 16,881 shapes categorized into 16 classes, with 50 part labels. Within each class, the number of parts varies from 2 to 6. As described in Sec.~\ref{sub_sec:shapenet_dataset}, we aim to augment the challenge of the ShapeNetPart dataset and enhance its representation of real-world point cloud scenarios. To achieve this, we apply various random operations to the dataset, introducing noise to simulate imperfect point clouds commonly encountered in real-world environments. 
Then, we transform all part segmentation labels of a sample into points in space, each point containing XYZ coordinates. These transformed points are then aggregated to construct a point cloud with the same shape as the input. Finally, we combine the input with the labeled point cloud to create a sample in our dataset.

\begin{figure*}[t]
\centering
\includegraphics[width=0.99\textwidth]{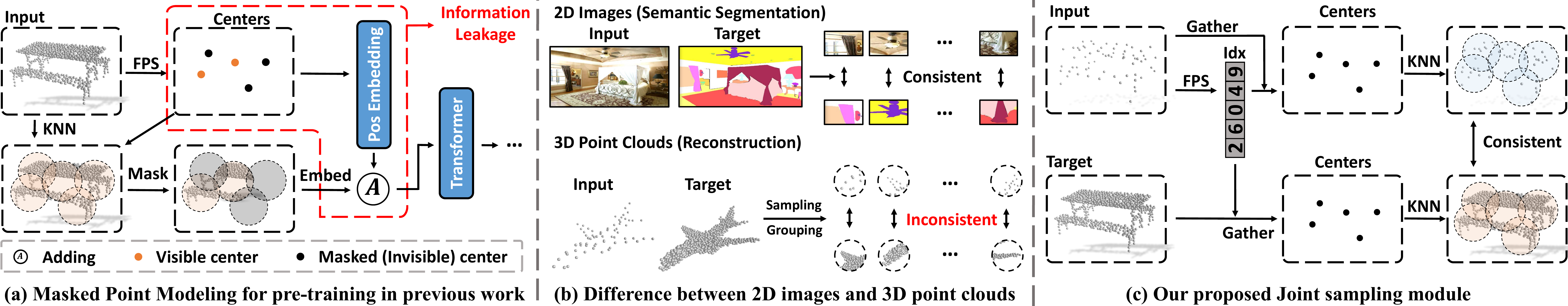}
\caption{
\textbf{(a) The pre-training pipeline used in previous works.} When performing Masked Point Modeling (MPM), these works~\cite{pointbert,pointmae,pointm2ae} use the center position of the target patches for position embedding, which results in information leakage. 
\textbf{(b) Difference between 2D images and 3D point clouds.} For 2D images, the semantic information at corresponding pixel positions in the input and target images is consistent. However, after performing grouping operations on 3D point clouds, the order of points will inevitably be disrupted, resulting in inconsistency between the point sequences of the input and output point clouds. 
\textbf{(c) Joint Sampling module} involves recording the indices of sampled center points and employing the K-nearest neighbor strategy to sample both the input and target point clouds concurrently.
}
\label{fig:technical_contri}
\end{figure*}

\noindent
\textbf{Human3D}~\cite{human3d} is a dataset comprising point cloud videos captured in real-world scenarios, annotated based on human body parts. 
Each sample of it contains two persons, both divided into 15 parts.
To seamlessly integrate prompts into our proposed in-context learning framework, we divide each sample containing two individuals into two separate samples. Each sample focuses on one person with annotations for human-part segmentation.
Given the complexity and difficulty in distinguishing between the original 15 body parts, we merge them into 6 categories: head, body, left arm, right arm, left leg, and right leg. This consolidation aims to enhance the semantic representation of human body parts. Additionally, we treat each frame of the dynamic video as a static sample in our dataset. This approach allows us to create a single-person human body part segmentation dataset, 
comprising 57,078 samples.

\noindent
\textbf{BEHAVE}~\cite{behave} is an extensive dataset focusing on full-body human-object interactions, featuring multi-view RGBD frames paired with corresponding 3D SMPL and object fits. Additionally, it includes annotated contact points between humans and objects. The BEHAVE dataset encompasses roughly 15K frames capturing 8 subjects engaging in various interactions with 20 commonly encountered objects.
Presented in video format, each frame of the RGBD video is transformed into point clouds, and treated as an individual sample within our dataset. These point cloud samples are annotated for segmentation, covering both the person and the object. 
After gathering around 15k point cloud samples, we excluded samples in which the proportion of points attributed to objects is smaller than 5\% of the entire point cloud sample.

\noindent
\textbf{AKB-48}~\cite{akb48} is a comprehensive-articulated object dataset containing 2,037 real-world 3D models across 48 categories. Each model is meticulously scanned from its real-life counterpart and further refined manually. Throughout the data collection, AKB-48 categorizes objects into multiple components, facilitating their utilization in part segmentation tasks.
However, some categories in AKB-48 inherently pose challenges due to semantic ambiguity or confusion. 
For example, within the ``toy'' category, samples may include both dinosaurs and cars, which do not share identical labels, highlighting semantic ambiguity.
Another instance is the ``folding rack'' category, which presents a symmetric object segmented into two or even four components despite its identical left and right sides, showcasing semantic confusion. 
To address these issues, we conduct filtering and preprocessing operations on the AKB-48 dataset. 
Firstly, categories prone to semantic confusion are removed. Furthermore, we manually consolidate components with similar semantics into a unified category. 
Finally, following a similar approach to the ShapeNetPart dataset, each component is labeled and merged to form a comprehensive point cloud with segmentation labels. 
Given AKB-48's diverse component categories and absence of overlap with previous datasets, we consider it a one-shot segmentation generalization test dataset.

\section{Methodology}
\label{sec:methodology}

In this section, we first introduce our Point-In-Context (PIC), a general point cloud framework capable of handling multiple tasks via the in-context learning capability. 
Next, to enhance the performance and generalization of PIC in segmentation tasks, we highlight the extended version of PIC, termed PIC++, which not only retains the multitask capabilities but also excels in part segmentation across multiple datasets.

\begin{figure*}[t]
\hsize=\textwidth
\centering
\includegraphics[width=\textwidth]{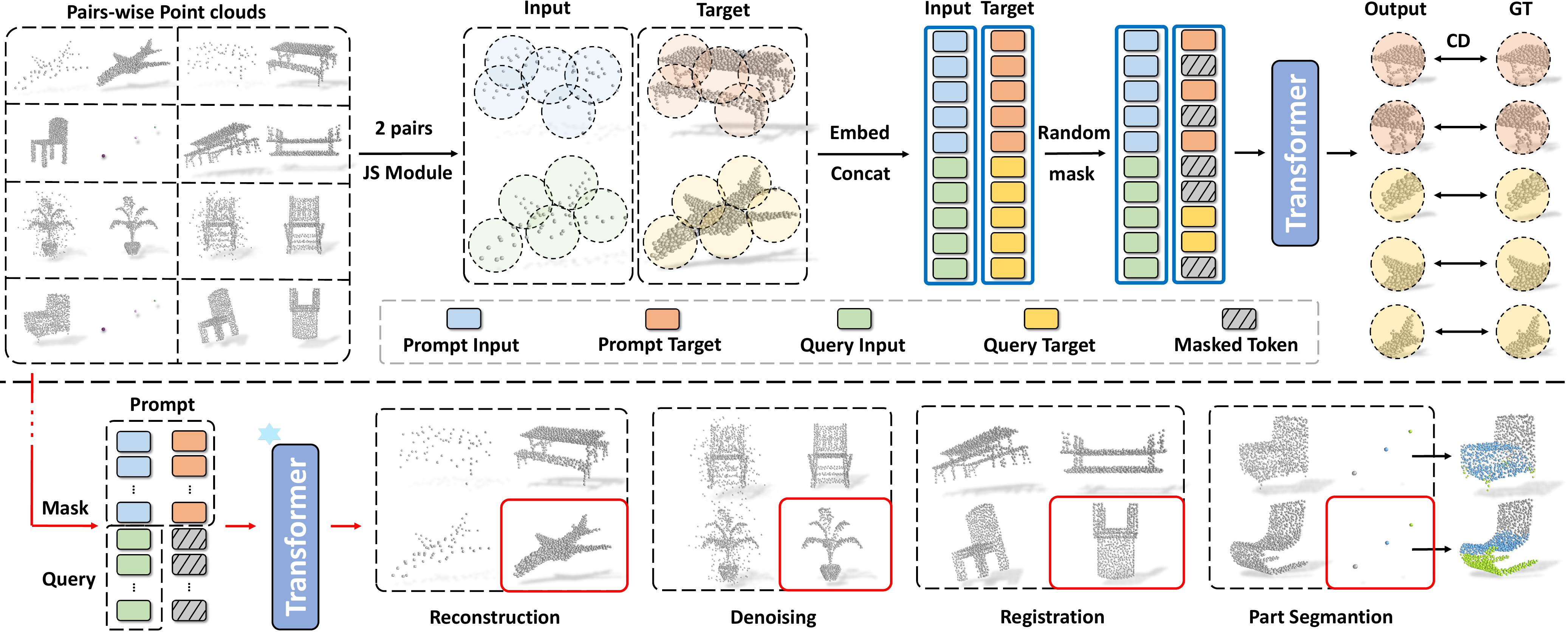}
\caption{
\textbf{Overall scheme of our Point-In-Context~(PIC)}.
\textit{Top}: Training pipeline of the Masked Point Modeling~(MPM) framework. During training, each sample comprises two pairs of input and target point clouds that tackle the same task. These pairs are fed into the transformer model to perform the masked point reconstruction task, which follows a random masking process. \textit{Bottom}: In-context inference on multitask. Our PIC could infer results on various downstream point cloud tasks, including reconstruction, denoising, registration, and part segmentation. 
}
\label{fig:framework}
\end{figure*}

\subsection{Point-In-Context Model}
\label{sub_sec:pic_model}

\noindent
\textbf{MPM for 3D In-Context Learning.}
Building upon previous works~\cite{visualprompt,painter}, we adopt Masked Point Modeling~(MPM) for the training of point cloud understanding and introduce Point-In-Context (PIC). 
In PIC, we conceptualize the local area of point clouds within the MPM as akin to image tokens within the MIM, thereby enabling us to harness the power of the transformer architecture for processing both point cloud and image data seamlessly. 
Illustrated in Fig.~\ref{fig:framework}, our training process aims to reconstruct the masked point patches, facilitated by the presence of input point clouds and visible target point cloud patches. 
During the inference stage, given a query point cloud, PIC will perform the specific task: reconstruction, denoising, registration, or part segmentation, according to the prompt. 
Subsequently, it generates target outputs corresponding to the query according to the task demonstrated by the prompt.

\noindent
\textbf{Information Leakage.} 
The conventional 3D MPM approach of employing position embeddings may introduce information leakage. That is because previous pre-training pipelines embed positional information with the sampled center point coordinates of all patches, even for those that are masked out (invisible), as shown in Fig.~\ref{fig:technical_contri}~(a). 
The previous works~\cite{pointmae,pointm2ae} treat the MPM framework as a pre-training strategy, followed by fine-tuning on specific downstream tasks, such as classification or segmentation. In contrast, our PIC is an end-to-end model and does not adhere to the ``pre-training then fine-tuning'' training paradigm. Therefore, if we continue to use the center points masked out from the target point clouds as the input for positional embedding while using the MPM framework, we will not be able to provide the coordinates of the center points of the target point clouds for this positional embedding during inference, or it will lead to information leakage.

\begin{figure*}[t]
\centering
\includegraphics[width=0.99\textwidth]{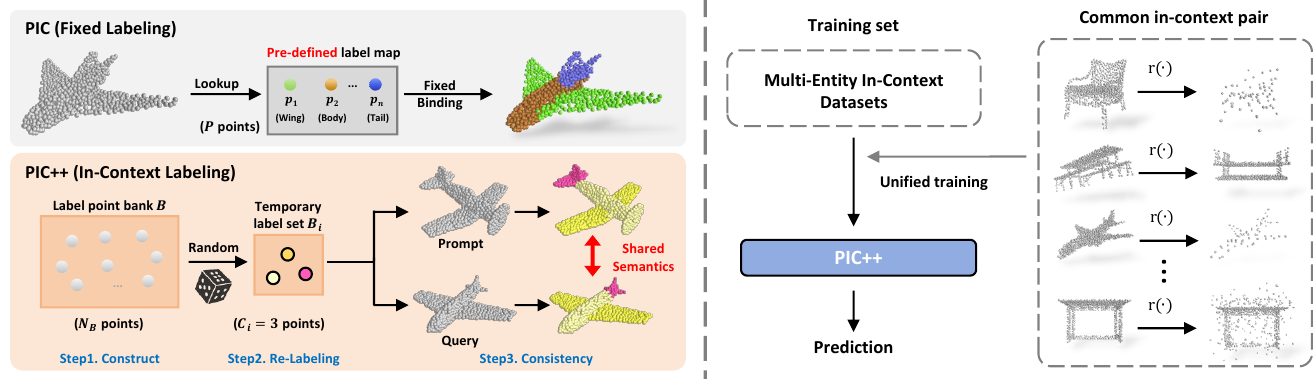}
\caption{
\textbf{(a) Comparison of generating targets between In-Context Labeling~(PIC++) and pre-defined label map~(PIC).} $P$ represents the number of all parts~($P\gg N_{B} > |C_i|$). \textbf{(b) Illustration of In-Context Enhancing.} We first randomly apply common corruptions on a clean point cloud and construct in-context pairs. Then we train our PIC++ on the combined training set of Multi-Entity In-Context Datasets and the constructed in-context pairs.
}
\label{fig:in-context_labeling}
\vspace{-1em}
\end{figure*}

\noindent
\textbf{Joint Sampling (JS) Module.} 
Unlike pixels in 2D images that are aligned on the grid, 3D points in the point cloud are scattered and unordered~\cite{pointnet}, as illustrated in Fig~\ref{fig:technical_contri}~(b). Therefore, direct sampling or grouping based on location will lead to inconsistent results over input and target.
The operations can become meaningful only when the correspondence between input and target point clouds can be established. To achieve this, they must operate on the same set of point indices across inputs and targets instead of applying them to the same spatial region.
To handle the above issues, we collect $N_{C}$ central points from each input point cloud and retrieve their indices, which we then use to obtain the center points of every patch in both the input and target point clouds. The process is shown in Fig.~\ref{fig:technical_contri}~(c). 
The key of our JS module is the consistency between center point indices of corresponding patches in both target and input point clouds. In other words, the order of the input token sequence and the target token sequence are well-aligned after being processed by the JS module. Such a design compensates for the missing positional embedding of the target while avoiding information leakage. Therefore, it facilitates the model in learning the inherent association between inputs and targets and streamlines the learning process. Subsequently, all point clouds search for neighborhoods containing $M$ points based on the center points corresponding to each patch.

\noindent
\textbf{Model Architecture.}
We use a standard transformer with an encoder-decoder structure as the backbone of our Point-In-Context, and a simple $1 \times 1$ convolutional layer as the task head for point cloud reconstruction. Inspired by Painter~\cite{painter} and MAE~\cite{mae}, we explore two different baselines for PIC, naming them PIC-Sep and PIC-Cat. For PIC-Sep, we take the input and masked target point clouds parallel to the transformer and then merge their features after several blocks, using a simple average operation. For PIC-Cat, we concatenate the input and target to form a new point cloud. Then, we mask it globally and feed it to the transformer for prediction. We denote the prompt pair as $R^k_i =\{P_i, T^k_i \}$ and query inputs,  $Q^k_j = \{ P_j, T^k_j\} $, then PIC-Sep and PIC-Cat can be formalized as:
\begin{equation}
    \resizebox{.9\hsize}{!}{$P^{\text{Sep}} = \text{Transformer}([P_i\parallel P_j], ([T^k_i\parallel T^k_j], M)),$}
\end{equation}
\begin{equation}
    \resizebox{.9\hsize}{!}{$P^{\text{Cat}} = \text{Transformer}([P_i\parallel T^k_i\parallel P_j\parallel  T^k_j], M),$}
\end{equation}
where $\parallel $ is the concatenate operation, and $M$ is the masked token to replace the invisible token.

\noindent
\textbf{Loss Function.}
The model is trained to reconstruct the masked point patches. To this end, we use the $\ell_{2}$ Chamfer Distance~\cite{chamferdistance}~(CD) as the training loss. Specifically, we calculate the Chamfer Distance between each predicted patch $P$ and its corresponding ground truth $G$, which can be formulated as: 

\begin{equation}
\begin{split}
    \mathcal{L}_{CD}(P, G) = \frac{1}{\left | P \right |}\sum_{p \in P} \min_{g\in G}\|p-g\|_2^2 + \\ 
    \frac{1}{\left | G \right |}\sum_{g\in G} \min_{p\in P}\|p-g\|_2^2.
\end{split}
\end{equation}

\subsection{Point-In-Context++ Model}
\label{sub_sec:sic_model}
Although PIC successfully demonstrates multitasking capabilities through the MPM framework with the JS module, it still has limitations in segmentation tasks. Therefore, to empower the segmentation capability of PIC, we further propose another version of PIC called PIC++, comprising two primary novel training strategies: In-Context Labeling and In-Context Enhancing.

\begin{algorithm*}[h]
\caption{In-Context Labeling: Construction and Dynamic Assignment}
\label{alg:icl_process}
\begin{algorithmic}[1]
\Require Label Bank Size $N_B$, Point Cloud Dataset $\mathcal{D}$
\Ensure Label Point Bank $B$, Mapped Labels for Batch

\State \textbf{Stage 1: Constructing the Label Point Bank}
\State Initialize empty bank $B \leftarrow \emptyset$
\For{$k = 1$ to $N_B$}
    \State Sample $p_k \sim \text{Uniform}([0, 1]^3)$ // \textit{Sample from unit cube}
    \State $B \leftarrow B \cup \{p_k\}$
\EndFor

\State
\State \textbf{Stage 2: Dynamic Re-labeling (Per Iteration)}
\While{Training or Inference}
    \State Get current object category $C$ and its semantic parts list $L_C = \{l_1, l_2, \dots, l_m\}$
    \State // \textit{Randomly select subset from Bank}
    \State Sample indices $I_{sub} \subset \{1, \dots, N_B\}$ such that $|I_{sub}| = m$
    \State Extract temporary label points $B_{sub} = \{B[i] \mid i \in I_{sub}\}$
    
    \State // \textit{Establish Mapping: Semantic Part $\to$ Coordinate}
    \State Initialize Mapping $\Phi \leftarrow \emptyset$
    \For{$j = 1$ to $m$}
        \State $\Phi(l_j) \leftarrow B_{sub}[j]$
    \EndFor
    
    \State // \textit{Apply Consistent Mapping to Prompt and Query}
    \State Prompt Labels $Y_{prompt} \leftarrow \{\Phi(y) \mid y \in \text{Prompt Part Categories}\}$
    \State Query Labels $Y_{query} \leftarrow \{\Phi(y) \mid y \in \text{Query Part Categories}\}$
    
    \State Proceed with Model Optimization/Inference using $Y_{prompt}$ and $Y_{query}$
\EndWhile
\end{algorithmic}
\end{algorithm*}

\noindent
\textbf{Limitations of PIC on Segmentation Tasks.}
In the PIC framework, to adjust to the input-output formats of other point cloud tasks, all part labels in the segmentation task are converted into points with XYZ coordinates in space, forming a pre-defined label point map. 
The label point map is fixed during training, leading to two main issues: 
\textbf{First}, fixed label points make the model rely solely on the static three-dimensional positions of label points for point cloud segmentation.
This is not conducive to learning the mapping relationship in example pairs (prompts), contradicting the original intention of in-context learning. Therefore, part segmentation cannot gain mutual benefit from other sub-tasks.
\textbf{Second}, when additional part segmentation datasets are incorporated, the large set of label points crowd in the limited space, exacerbating the model's difficulty in regressing prediction to the correct partition. 
As a result, a small regression error has a higher possibility of leading to a wrong prediction.
\textbf{Third}, once the label-point assignment is fixed, the space partition is determined that each label covers a specific region. 
In this case, allocating new space to novel classes is challenging. Therefore, the fixed label choice sacrifices the possibility of generalization.

\noindent
\textbf{In-Context Labeling.} To address the limitations of the static assignment in PIC, we propose a dynamic strategy named In-Context Labeling (ICL). Unlike previous methods that rely on a fixed one-to-one binding between label points and semantic classes, ICL employs a dynamic, context-aware assignment mechanism. The process consists of three key steps, as illustrated in Fig. \ref{fig:in-context_labeling} and Algorithm.~\ref{alg:icl_process}:

\textbf{Constructing the Label Point Bank.}
First, we initialize a comprehensive label point bank $B = \{p_k \in \mathbb{R}^3, k=1,2,\cdots,N_B\}$ by uniformly sampling the 3D coordinate space. The size of the bank, $N_B$, is set to satisfy:
\begin{equation}
N_B \geq \max_{i \leq n} |C_i|,
\end{equation}
where $C_i$ denotes the number of semantic categories in the $i$-th sample, and $n$ represents the total dataset size.

\textbf{Dynamic Re-labeling.}
The second step involves dynamically assigning labels for each training iteration. In contrast to traditional methods where a specific spatial point permanently represents a specific component (e.g., a fixed point for ``wing"), we break this static binding. For each input point cloud $i$, we randomly sample $|C_i|$ points from the bank $B$ to construct a temporary label set $B_i$:
\begin{equation}
B_{i} \subset B, \quad \text{s.t.} \quad |B_{i}| = |C_i|.
\end{equation}
These selected points are then randomly mapped to the semantic parts of the current point cloud. This stochastic assignment forces the model to decouple semantic meaning from absolute spatial coordinates, thereby facilitating the learning of novel categories.

\textbf{Prompt Consistency.}
Finally, to enable in-context learning, we select a prompt $R_i$ (support sample) that shares the same semantic-to-label mapping as the query input $Q_j$ from the training set. This consistency is crucial; it serves as a contextual anchor. By observing how the label points in $B_i$ are applied to the prompt $R_i$, the model infers the segmentation rule for the current query.

\textbf{Benefits.}
Our context-based labeling approach allows the model to focus more on the semantic information presented in the prompts. Besides, it enables our framework to easily incorporate additional segmentation datasets without increasing the complexity of identifying point cloud semantic information. 
During training, the model also segments the query point cloud by observing the hidden pattern between the input point cloud and the label point in the prompt rather than relying on the label point's three-dimensional coordinates.
Thus, the in-context labeling provides the scalability and extensibility for the model to generalize to other segmentation datasets. We only need to randomly assign label points to the point clouds in out-of-domain segmentation datasets and select a random sample from the same category as the prompt. This ensures the prompt and query share consistent label definitions, a process that is fully automated without handcrafted intervention.

\begin{table*}[t]
\caption{\textbf{Comparison results on ShapeNet In-Context Datasets}, including reconstruction, denoising, registration, and part segmentation. PIC and PIC++ represent models described in Sec.\ref{sub_sec:pic_model} and Sec.\ref{sub_sec:sic_model} respectively.
Our PICs only need to be trained once and do not require fine-tuning or any parameter updates during testing. We also report task-specific and multitask models. $\dagger$ indicates models with a shared backbone and multiple task heads. $\ddagger$ indicates models whose backbone is pre-trained via specific tasks. For reconstruction, denoising, and registration, we report the Chamfer Distance~\cite{chamferdistance}~(CD) loss ($\times$1000). We report the mean Intersection over Union (mIoU) metric for part segmentation. Bold: best. Underline: second-best. \textcolor{red}{red} and \textcolor{blue}{blue} arrows indicate whether PIC++ is worse or better than PIC respectively.} 
\setlength\tabcolsep{1mm}
\renewcommand\arraystretch{1.2}
\centering
\scalebox{0.7}{
\begin{tabular}{llcccccc!{\vrule width 1pt}cccccc!{\vrule width 1pt}cccccc!{\vrule width 1pt}cc}
\toprule
\multicolumn{2}{c!{\vrule width 1pt}}{\multirow{2}{*}{Models}} & \multicolumn{1}{c!{\vrule width 1pt}}{\multirow{2}{*}{Venues}} & \multicolumn{6}{c}{Reconstruction CD $\downarrow$}         & \multicolumn{6}{c}{Denoising CD $\downarrow$}           & \multicolumn{6}{c}{Registration CD $\downarrow$}     & Part Seg.  \\
\multicolumn{2}{c!{\vrule width 1pt}}{}    & \multicolumn{1}{c!{\vrule width 1pt}}{}                   & L1    & L2    & L3    & L4    & L5    & \multicolumn{1}{c!{\vrule width 1pt}}{Avg.}  & L1    & L2    & L3    & L4    & L5    & \multicolumn{1}{c!{\vrule width 1pt}}{Avg.}  & L1    & L2    & L3    & L4    & L5    & \multicolumn{1}{c!{\vrule width 1pt}}{Avg.}  & mIoU$\uparrow$              \\ \hline
\rowcolor{gray!10}
\multicolumn{22}{c}{Task-Specific Models: trained separately for single task}           \\ \hline
\rowcolor{gray!10}
\multicolumn{2}{l!{\vrule width 1pt}}{PointNet~\cite{pointnet}} & \multicolumn{1}{c!{\vrule width 1pt}}{CVPR'17}     & 3.7   & 3.7   & 3.8   & 3.9   & 4.1   & \multicolumn{1}{c!{\vrule width 1pt}}{3.9}   & 4.1   & 4.0   & 4.1   & 4.0   & 4.2   & \multicolumn{1}{c!{\vrule width 1pt}}{4.1}   & 5.3   & 5.9   & 6.9   & 7.7   & 8.5   & \multicolumn{1}{c!{\vrule width 1pt}}{6.9}   & 77.45      \\
\rowcolor{gray!10}
\multicolumn{2}{l!{\vrule width 1pt}}{DGCNN~\cite{dgcnn}}  & \multicolumn{1}{c!{\vrule width 1pt}}{TOG'19}   & 3.9   & 3.9   & 4.0   & 4.1   & 4.3   & \multicolumn{1}{c!{\vrule width 1pt}}{4.0}   & 4.7   & 4.5   & 4.6   & 4.5   & 4.7   & \multicolumn{1}{c!{\vrule width 1pt}}{4.6}   & 6.2   & 6.7   & 7.3   & 7.4   & 7.7   & \multicolumn{1}{c!{\vrule width 1pt}}{7.1}   & 76.12   \\
\rowcolor{gray!10}
\multicolumn{2}{l!{\vrule width 1pt}}{PCT~\cite{pct}}  & \multicolumn{1}{c!{\vrule width 1pt}}{CVM'21}  & 2.4   & 2.4   & 2.5   & 2.6   & 3.0   & \multicolumn{1}{c!{\vrule width 1pt}}{2.6}   & 2.3   & 2.2   & 2.2   & 2.2   & 2.3   & \multicolumn{1}{c!{\vrule width 1pt}}{2.2}   & 5.3   & 5.7   & 6.3   & 6.9   & 7.2   & \multicolumn{1}{c!{\vrule width 1pt}}{6.3}   & 79.46      \\
\rowcolor{gray!10}
\multicolumn{2}{l!{\vrule width 1pt}}{ACT~\cite{act}}  & \multicolumn{1}{c!{\vrule width 1pt}}{ICLR'23}  & 2.4   & 2.5   & 2.3   & 2.5   & 2.8   & \multicolumn{1}{c!{\vrule width 1pt}}{2.5}   & 2.2   & 2.3   & 2.2   & 2.3   & 2.5   & \multicolumn{1}{c!{\vrule width 1pt}}{2.3}   & 5.1   & 5.6   & 5.9   & 6.0   & 7.0   & \multicolumn{1}{c!{\vrule width 1pt}}{5.9}   & 81.24      \\ \hline
\multicolumn{22}{c}{Multitask Models~$^{\dagger}$: shared backbone + multiple task heads}               \\ \hline
\multicolumn{2}{l!{\vrule width 1pt}}{PointNet$^{\dagger}$~\cite{pointnet}}   & \multicolumn{1}{c!{\vrule width 1pt}}{CVPR'17}   & 87.2  & 86.6  & 87.3  & 90.8  & 92.2  & \multicolumn{1}{c!{\vrule width 1pt}}{88.8}  & 17.8  & 22.0  & 25.6  & 30.4  & 33.2  & \multicolumn{1}{c!{\vrule width 1pt}}{25.8}  & 25.4  & 22.6  & 24.9  & 25.7  & 26.9  & \multicolumn{1}{c!{\vrule width 1pt}}{25.1}  & 15.33      \\
\multicolumn{2}{l!{\vrule width 1pt}}{DGCNN$^{\dagger}$~\cite{dgcnn}}  & \multicolumn{1}{c!{\vrule width 1pt}}{TOG'19}   & 38.8  & 36.6  & 37.5  & 37.9  & 42.9  & \multicolumn{1}{c!{\vrule width 1pt}}{37.7}  & 6.5   & 6.3   & 6.5   & 6.4   & 7.1   & \multicolumn{1}{c!{\vrule width 1pt}}{6.5}   & 12.5  & 14.9  & 17.9  & 19.7  & 20.7  & \multicolumn{1}{c!{\vrule width 1pt}}{17.1}  & 16.95      \\
\multicolumn{2}{l!{\vrule width 1pt}}{PCT$^{\dagger}$~\cite{pct}}  & \multicolumn{1}{c!{\vrule width 1pt}}{CVM'21}   & 34.7  & 44.1  & 49.9  & 50.0  & 52.3  & \multicolumn{1}{c!{\vrule width 1pt}}{46.2}  & 11.2  & 10.3  & 10.7  & 10.2  & 10.5  & \multicolumn{1}{c!{\vrule width 1pt}}{10.6}  & 24.4  & 26.0  & 29.6  & 32.8  & 34.7  & \multicolumn{1}{c!{\vrule width 1pt}}{29.5}  & 16.71      \\ \hline
\multicolumn{22}{c}{Multitask Models~$^{\ddagger}$: pre-trained shared backbone + multiple task heads}    \\ \hline
\multicolumn{2}{l!{\vrule width 1pt}}{PointNet$^{\ddagger}$~\cite{pointnet}}   & \multicolumn{1}{c!{\vrule width 1pt}}{CVPR'17}   & 47.0  & 45.8  & 45.4  & 45.4  & 45.8  & \multicolumn{1}{c!{\vrule width 1pt}}{45.9}  & 22.9  & 23.2  & 26.3  & 28.3  & 30.0  & \multicolumn{1}{c!{\vrule width 1pt}}{26.1}  & 35.5  & 34.8  & 37.1  & 37.2  & 38.6  & \multicolumn{1}{c!{\vrule width 1pt}}{36.6}  & 10.13      \\
\multicolumn{2}{l!{\vrule width 1pt}}{DGCNN$^{\ddagger}$~\cite{dgcnn}}  & \multicolumn{1}{c!{\vrule width 1pt}}{TOG'19}   & 46.7   & 47.2  & 48.1  & 48.6  & 48.5  & \multicolumn{1}{c!{\vrule width 1pt}}{47.8}  & 8.2   & 8.3   & 8.4   & 8.8   & 9.2   & \multicolumn{1}{c!{\vrule width 1pt}}{8.6}   & 14.2  & 15.8  & 18.2  & 21.8  & 23.5  & \multicolumn{1}{c!{\vrule width 1pt}}{18.7}  &  21.35     \\
\multicolumn{2}{l!{\vrule width 1pt}}{PCT$^{\ddagger}$~\cite{pct}}  & \multicolumn{1}{c!{\vrule width 1pt}}{CVM'21}   & 64.7  & 60.8  & 59.2  & 60.1  & 59.7  & \multicolumn{1}{c!{\vrule width 1pt}}{61.0}  & 14.5  & 12.2  & 12.4  & 12.0  & 11.8  & \multicolumn{1}{c!{\vrule width 1pt}}{12.6}  & 22.6  & 25.2  & 28.3  & 31.1  & 33.2  & \multicolumn{1}{c!{\vrule width 1pt}}{28.1}  & 15.43      \\
\multicolumn{2}{l!{\vrule width 1pt}}{PointMAE$^{\ddagger}$~\cite{pointmae}} & \multicolumn{1}{c!{\vrule width 1pt}}{ECCV'22}  & 5.5  & 5.5  & 6.1  & 6.4  & 6.4  & \multicolumn{1}{c!{\vrule width 1pt}}{6.0}  & 5.6  & 5.4  & 5.6  & 5.5  & 5.8  & \multicolumn{1}{c!{\vrule width 1pt}}{5.6}  & 11.4  & 12.8  & 14.8  & 16.0  & 16.9  & \multicolumn{1}{c!{\vrule width 1pt}}{14.5}  & 5.42      \\ 
\multicolumn{2}{l!{\vrule width 1pt}}{ACT$^{\ddagger}$~\cite{act}} & \multicolumn{1}{c!{\vrule width 1pt}}{ICLR'23}  & 7.4  & 6.6  & 6.5  & 6.6  & 7.0  & \multicolumn{1}{c!{\vrule width 1pt}}{6.8}  & 7.3  & 6.8  & 7.0  & 6.8  & 7.2  & \multicolumn{1}{c!{\vrule width 1pt}}{7.0}  & 12.2  & 14.4  & 19.4  & 25.5  & 29.0  & \multicolumn{1}{c!{\vrule width 1pt}}{20.1}  & 12.08      \\ 
\multicolumn{2}{l!{\vrule width 1pt}}{I2P-MAE$^{\ddagger}$~\cite{i2pmae}} & \multicolumn{1}{c!{\vrule width 1pt}}{CVPR'23}  & 17.0  & 16.0  & 16.7  & 17.2  & 18.5  & \multicolumn{1}{c!{\vrule width 1pt}}{17.2}  & 20.6  & 20.4  & 20.1  & 18.3  & 18.8  & \multicolumn{1}{c!{\vrule width 1pt}}{19.6}  & 32.5  & 31.3  & 31.1  & 31.6  & 31.2  & \multicolumn{1}{c!{\vrule width 1pt}}{31.5}  & 22.60      \\ 
\multicolumn{2}{l!{\vrule width 1pt}}{ReCon$^{\ddagger}$~\cite{recon}} & \multicolumn{1}{c!{\vrule width 1pt}}{ICML'23}  & 12.4  & 12.1  & 12.4  & 12.5  & 13.1  & \multicolumn{1}{c!{\vrule width 1pt}}{12.5}  & 20.4  & 24.5  & 27.2  & 29.2  & 32.5  & \multicolumn{1}{c!{\vrule width 1pt}}{26.9}  & 14.7  & 16.3  & 19.2  & 21.5  & 22.5  & \multicolumn{1}{c!{\vrule width 1pt}}{18.8}  & 7.71      \\
\multicolumn{2}{l!{\vrule width 1pt}}{MAMBA3D$^{\ddagger}$~\cite{han2024mamba3d}} & \multicolumn{1}{c!{\vrule width 1pt}}{MM'24}  & 8.4 & 7.8 & 9.3 & 9.8 & 10.5 & \multicolumn{1}{c!{\vrule width 1pt}}{9.2} & 13.2 & 13.8 & 13.5 & 14.8 & 12.8 & \multicolumn{1}{c!{\vrule width 1pt}}{13.6} & 9.9 & 13.2 & 13.9 & 19.7 & 20.3 & \multicolumn{1}{c!{\vrule width 1pt}}{15.4} & 14.58 \\
\multicolumn{2}{l!{\vrule width 1pt}}{PCMamba$^{\ddagger}$~\cite{pcm}} & \multicolumn{1}{c!{\vrule width 1pt}}{AAAI'25}  & 8.1 & 8.9 & 10.2 & 10.5 & 11.7 & \multicolumn{1}{c!{\vrule width 1pt}}{9.9} & 14.5 & 14.2 & 15.9 & 14.7 & 14.5 & \multicolumn{1}{c!{\vrule width 1pt}}{14.8} & 11.1 & 13.8 & 18.2 & 21.8 & 22.7 & \multicolumn{1}{c!{\vrule width 1pt}}{17.5} & 19.53 \\ \hline
\multicolumn{22}{c}{In-context Learning Models: task-agnostic architecture for multiple tasks}                     \\ \hline
\multicolumn{2}{l!{\vrule width 1pt}}{Copy}      & \multicolumn{1}{c!{\vrule width 1pt}}{/}              & 155 & 153 & 152 & 156 & 155 & \multicolumn{1}{c!{\vrule width 1pt}}{154} & 149 & 155 & 157 & 155 & 155 & \multicolumn{1}{c!{\vrule width 1pt}}{154} & 155 & 157 & 156 & 148 & 154 & \multicolumn{1}{c!{\vrule width 1pt}}{154} & 24.18    \\
\multicolumn{2}{l!{\vrule width 1pt}}{Point-BERT~\cite{pointbert}} & \multicolumn{1}{c!{\vrule width 1pt}}{CVPR'22}   & 288 & 285 & 292 & 286 & 308 & \multicolumn{1}{c!{\vrule width 1pt}}{292} & 292 & 293 & 298 & 296 & 299 & \multicolumn{1}{c!{\vrule width 1pt}}{296} & 291 & 295 & 294 & 295 & 298 & \multicolumn{1}{c!{\vrule width 1pt}}{294} & 0.65        \\ \hline
\multirow{2}{*}{"-Cat"} & \multicolumn{1}{l!{\vrule width 1pt}}{\textbf{PIC}}   & \multicolumn{1}{c!{\vrule width 1pt}}{NeurIPS'23}    & 3.2   & 3.6   & 4.6   & 4.9   & 5.5   & \multicolumn{1}{c!{\vrule width 1pt}}{\textbf{4.3}}   & 3.9   & 4.6   & 5.3   & 6.0   & 6.8   & \multicolumn{1}{c!{\vrule width 1pt}}{\underline{5.3}}   & 10.0  & 11.4  & 13.8  & 16.9  & 18.6  & \multicolumn{1}{c!{\vrule width 1pt}}{14.1}  & 78.95       \\
& \multicolumn{1}{l!{\vrule width 1pt}}{\textbf{PIC++}}   & \multicolumn{1}{c!{\vrule width 1pt}}{/}    & 4.5\textbf{\textcolor{red}{$^{\uparrow}$}}  & 3.3\textbf{\textcolor{blue}{$^{\downarrow}$}}  & 3.7\textbf{\textcolor{blue}{$^{\downarrow}$}}  & 4.6\textbf{\textcolor{blue}{$^{\downarrow}$}}  & 5.4\textbf{\textcolor{blue}{$^{\downarrow}$}}  & \multicolumn{1}{c!{\vrule width 1pt}}{\textbf{4.3$^{-}$}}  & 3.8\textbf{\textcolor{blue}{$^{\downarrow}$}}  & 4.3\textbf{\textcolor{blue}{$^{\downarrow}$}} & 5.1\textbf{\textcolor{blue}{$^{\downarrow}$}}  & 5.6\textbf{\textcolor{blue}{$^{\downarrow}$}}  & 6.7\textbf{\textcolor{blue}{$^{\downarrow}$}}  & \multicolumn{1}{c!{\vrule width 1pt}}{\textbf{5.1\textcolor{blue}{$^{\downarrow}$}}}  & 9.7\textbf{\textcolor{blue}{$^{\downarrow}$}}  & 11.6\textbf{\textcolor{red}{$^{\uparrow}$}}  & 12.8\textbf{\textcolor{blue}{$^{\downarrow}$}}  & 15.9\textbf{\textcolor{blue}{$^{\downarrow}$}}  & 18.0\textbf{\textcolor{blue}{$^{\downarrow}$}}  & \multicolumn{1}{c!{\vrule width 1pt}}{13.6\textbf{\textcolor{blue}{$^{\downarrow}$}}}  & \underline{85.27}\textbf{\textcolor{blue}{$^{\uparrow}$}}       \\ \hline
\multirow{2}{*}{"-Sep"} & \multicolumn{1}{l!{\vrule width 1pt}}{\textbf{PIC}}  & \multicolumn{1}{c!{\vrule width 1pt}}{NeurIPS'23}     & 4.7   & 4.3   & 4.3   & 4.4   & 5.7   & \multicolumn{1}{c!{\vrule width 1pt}}{4.7}   & 6.3   & 7.2   & 7.9   & 8.2   & 8.6   & \multicolumn{1}{c!{\vrule width 1pt}}{7.6}   & 8.6   & 9.2   & 10.2  & 11.3  & 12.4  & \multicolumn{1}{c!{\vrule width 1pt}}{\underline{10.3}}  & 74.95       \\ 
& \multicolumn{1}{l!{\vrule width 1pt}}{\textbf{PIC++}}  & \multicolumn{1}{c!{\vrule width 1pt}}{/}     & 4.5\textbf{\textcolor{blue}{$^{\downarrow}$}}  & 4.1\textbf{\textcolor{blue}{$^{\downarrow}$}}  & 4.2\textbf{\textcolor{blue}{$^{\downarrow}$}}   & 4.2\textbf{\textcolor{blue}{$^{\downarrow}$}}  & 5.9\textbf{\textcolor{red}{$^{\uparrow}$}}   & \multicolumn{1}{c!{\vrule width 1pt}}{4.6\textbf{\textcolor{blue}{$^{\downarrow}$}}}  & 5.8\textbf{\textcolor{blue}{$^{\downarrow}$}}  & 7.1\textbf{\textcolor{blue}{$^{\downarrow}$}}  & 7.7\textbf{\textcolor{blue}{$^{\downarrow}$}}  & 8.2\textbf{$^{-}$}   & 8.4\textbf{\textcolor{blue}{$^{\downarrow}$}}  & \multicolumn{1}{c!{\vrule width 1pt}}{7.4\textbf{\textcolor{blue}{$^{\downarrow}$}}}  & 7.6\textbf{\textcolor{blue}{$^{\downarrow}$}}  & 6.2\textbf{\textcolor{blue}{$^{\downarrow}$}}  & 6.8\textbf{\textcolor{blue}{$^{\downarrow}$}}  & 6.9\textbf{\textcolor{blue}{$^{\downarrow}$}}  & 7.9\textbf{\textcolor{blue}{$^{\downarrow}$}}  & \multicolumn{1}{c!{\vrule width 1pt}}{\textbf{7.1\textcolor{blue}{$^{\downarrow}$}}}  & \textbf{85.53\textcolor{blue}{$^{\uparrow}$}}       \\ 
\bottomrule
\end{tabular}
}
\label{tab:main_result_1}
\end{table*}

\textbf{In-Context Enhancing.}
As discussed above, we introduce a new training method of in-context labeling that involves completely randomly assigning labels to label points. This decoupling of semantics from fixed coordinates prevents the model from relying on rote memorization, but it also imposes a higher requirement on the model's ability to infer mapping rules from the context. To bridge this gap and further stimulate the in-context learning capability of the model, we introduce the In-Context Enhancing (ICE) training strategy.

Unlike static multi-task benchmarks (e.g., ShapeNet In-Context) that use fixed pre-defined tasks, we first dynamically apply random corruption operations to a clean point cloud $P_{j}$ and construct pairs of point clouds aiming to restore the damaged point cloud. The query point cloud pair can be formulated as:
\begin{equation}
    Q^{r}_{j}=\{r(P_{j}), P_{j}\},
\end{equation}
where $r(\cdotp )$ is a stochastic corruption function encompassing a diverse range of point cloud corruption operations (e.g., local masking, jittering, dropping). Each point cloud $P_{j}$ only contains XYZ coordinates without any color or component semantic information.

Additionally, we provide a pair of point clouds that have undergone the same corruption operation $r(\cdotp)$ as a prompt, requiring the model to restore the damaged point cloud based on the provided prompt. Subsequently, we involve the constructed pairs of point clouds in the training set to enhance the model's in-context learning capability during unified training (as illustrated in Fig.~\ref{fig:in-context_labeling}~(b)).

Critically, ICE serves as a geometric pretext task designed to synergize with In-Context Labeling. By forcing the model to identify varying degradation patterns in the prompt and transfer the restoration logic to the query, we foster a generalized ``meta-capability'' for analyzing prompt-query relationships. This differs fundamentally from standard multi-task learning, as the goal is not merely to solve the restoration task, but to train the model's sensitivity to contextual rules. Furthermore, by focusing solely on the XYZ coordinates of each point cloud without any additional color or semantic information, we ensure that the model learns to restore point clouds based on spatial context alone. This abstraction fosters a deeper understanding of the underlying structure of point clouds, enabling the model to learn more effectively within segmentation pairs across diverse datasets, which can be proved in Sec.~\ref{sub_sec:result_on_segmentation}.

\begin{figure*}[t]
\centering
\includegraphics[width=0.99\textwidth]{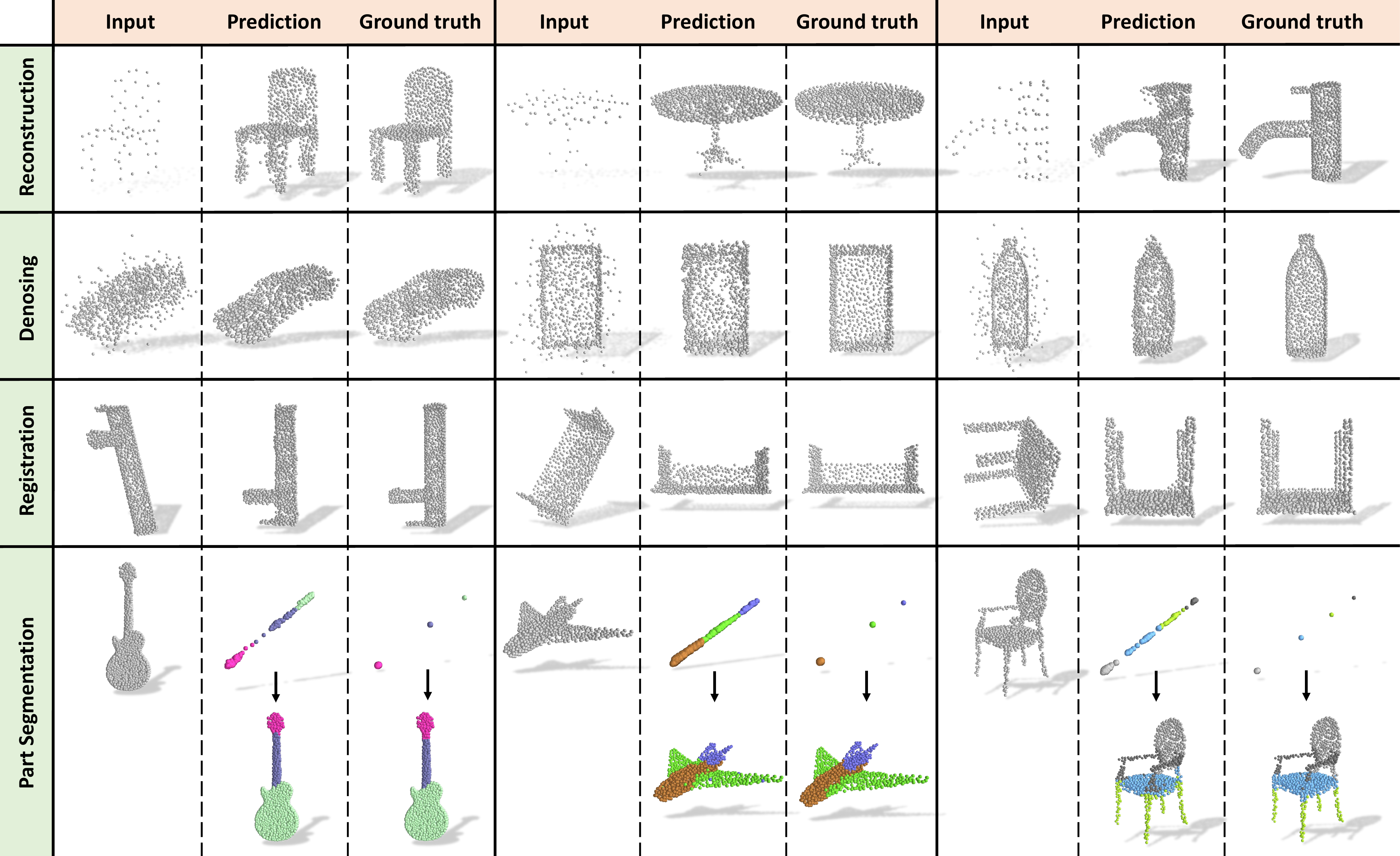}
\caption{
\textbf{Visualization of predictions from the "-Sep" variant of PIC on ShapeNet In-Context Datasets}. For part segmentation, we visualize the generated target together with the mapping back, both adding category-specific colors for a better look.
}
\label{fig:visualization_main}
\vspace{-1em}
\end{figure*}

\begin{figure*}[t]
\centering
\includegraphics[width=0.95\textwidth]{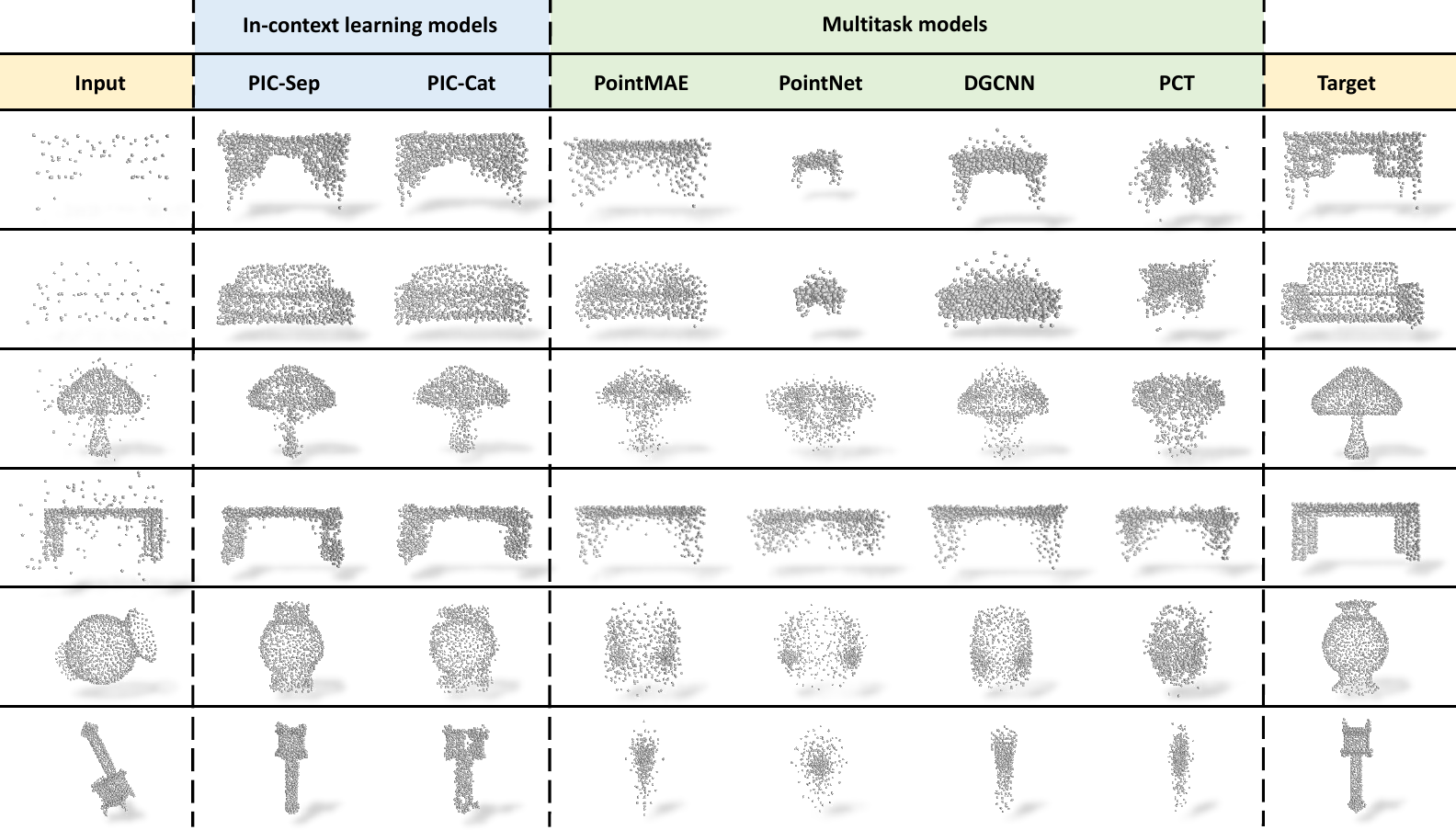}
\caption{
\textbf{Visualization of comparison results between PIC and multitask models} on reconstruction~(lines 1-2), denoising~(lines 3-4), and registration~(lines 5-6), where our models can generate more accurate predictions than other multitask models.
}
\vspace{-2em}
\label{fig:supp_comparison}
\end{figure*}

\begin{figure*}[t]
\centering
\includegraphics[width=0.9\textwidth]{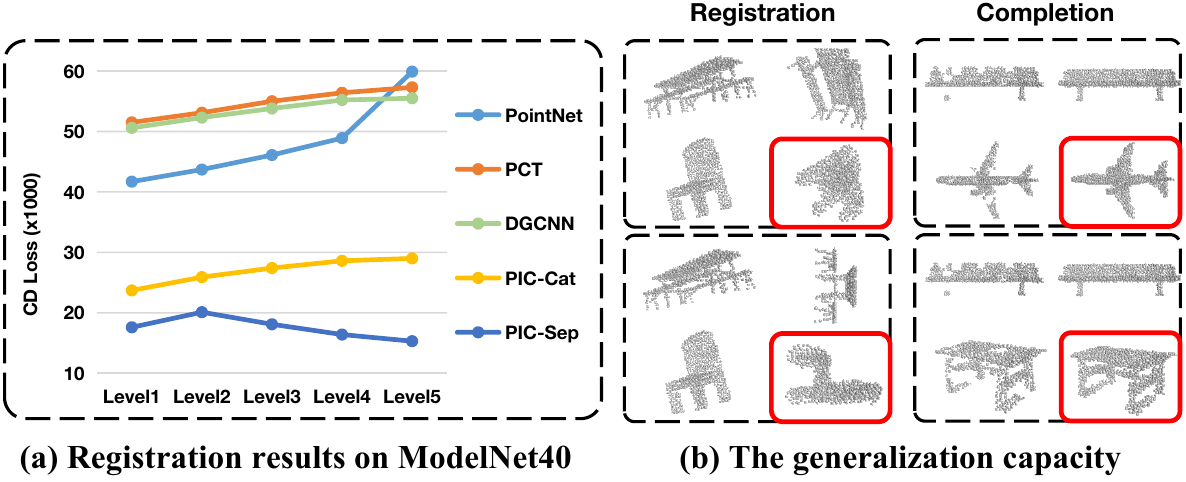}
\caption{
\textbf{(a) Generalized to out-of-distribution data.} We evaluate the registration task on the ModelNet40~\cite{modelnet40} dataset, which is not present in the training set. The Chamfer Distance (x1000) is used as the evaluation metric.
\textbf{(b) Generalized to new tasks.} Our model can rotate at any angle according to the example pairs and show the completion ability to the broken point cloud, which can be treated as an upgraded task of registration.
}
\vspace{-1em}
\label{fig:transfer_ability}
\end{figure*}

\noindent
\textbf{Inference of PIC++.}
In contrast to PIC's fixed segmentation target label point map, the in-context labeling training strategy introduces variability in the label point map for each point cloud. This difference poses unique challenges during evaluation, particularly when the target point cloud for the query is not visible. To address this issue, we leverage the label point bank $B_j$ utilized by the prompt $Q_j$ as the segmentation standard for the query.

During evaluation, for each point in the predicted point cloud, we compute its distance from each point in the label point set $B_j$. By selecting the label point with the minimum distance, we determine the segmentation result for each point. This process enables us to transform the dynamic regressed point cloud $\hat{P}_i$ into corresponding classification prediction while avoiding the fixed property from classification.

Although the dynamic assignment may introduce challenges to the model inference, this problem has been mitigated as the same mechanism is used for robust training. Therefore, no matter which label point set is randomly sampled, the model shall output consistent predictions. Secondly, dynamic assignment allows for the inclusion of novel classes during inference, embedding the model with the inherent generalization ability.
Overall, this methodology aligns with the principles of in-context learning, where the segmentation standard dynamically adjusts to the context of each query point cloud, thereby improving the model's generalization capabilities and segmentation accuracy~(see Sec.~\ref{sub_sec:result_on_segmentation}). 

\noindent
\textbf{Loss Function.}
As the Chamfer Distance~\cite{chamferdistance}~(CD) loss only measures the similarity between two point clouds, in our segmentation benchmark, we require precise regression of predictions for each point. Therefore, we incorporate Smooth-$\ell_{1}$~\cite{fastrcnn} loss to assist the model in optimization.
At the same time, our training purpose is to minimize the distance between the predicted point cloud and the target point cloud, so the CD loss in PIC is also used. The loss function of PIC++ can be formulated as:
\begin{equation}
    \mathcal{L}_{Seg}(P, G) = \mathcal{L}_{CD}(P, G) + \mathcal{L}_{Smooth-\ell_{1}}(P, G).
\end{equation}

\section{Experiments}
\label{sec:exp}

We first provide specific experimental details, including model parameters and training procedures, and information on other related methods~(Sec.~\ref{sub_sec:exp_details}). Subsequently, we quantitatively and qualitatively evaluate the proposed PIC and PIC++ models on the ShapeNet In-Context Dataset~(Sec.~\ref{sub_sec:result_on_multitasking}) and Multi-Entity In-Context Dataset~(Sec.~\ref{sub_sec:result_on_segmentation}), comparing them with other methods. The focus lies in exploring their performance in multitasking and part segmentation scenarios. Finally, we conduct detailed ablation studies to examine the rationality of the model architecture design and selection~(Sec.~\ref{sub_sec:ablation}).

\begin{table*}[t]
    \caption{\textbf{Influence of data scale.} We conduct experiments that include large-scale datasets Objaverse~\cite{objaverse} in the training set and analyze the impact of large-scale data on the model's in-context learning ability. The \textcolor{blue}{blue} arrow indicates the performance improvement brought by adding the Objaverse dataset.}
    \centering
    
    \begin{subtable}[t]{0.99\linewidth}
        \renewcommand\arraystretch{1.1}
        \setlength\tabcolsep{1.2mm}
        \caption{Experimental results of PIC on ShapeNet In-Context Datasets.}
        \centering
        \begin{tabular}{l|c|cccc}
            \toprule
            Variants & Objaverse     & Reconstruction CD$\downarrow$ & Denoising CD$\downarrow$ & Registration CD$\downarrow$ & Part Seg. mIoU$\uparrow$ \\ \hline
            \multirow{2}{*}{"-Cat"} & \XSolidBrush        & 4.3               & 5.3          & 14.1            & 78.95     \\
                        & \Checkmark       & 4.3$^{-}$    & 5.2\textcolor{blue}{$^{\downarrow 0.1}$}    & 10.7\textcolor{blue}{$^{\downarrow 3.4}$}            & 80.13\textcolor{blue}{$^{\uparrow 1.18}$}     \\ \hline
            \multirow{2}{*}{"-Sep"} & \XSolidBrush        & 4.7               & 7.6          & 10.3            & 74.95     \\
                        & \Checkmark       & 4.4\textcolor{blue}{$^{\downarrow 0.3}$}               & 6.2\textcolor{blue}{$^{\downarrow 1.4}$}          & 8.2\textcolor{blue}{$^{\downarrow 2.1}$}             & 81.96\textcolor{blue}{$^{\uparrow 7.01}$}     \\
            \bottomrule
            \end{tabular}
    \end{subtable}

    \begin{subtable}[t]{0.99\linewidth}
        \renewcommand\arraystretch{1.1}
        \setlength\tabcolsep{3.2mm}
        \caption{Generalization results of PIC on ModelNet40~\cite{modelnet40} on Registration. Note that ModelNet40 is not included in the training set. We report the CD Loss ($\times$1000).}
        \centering
        \begin{tabular}{l|c|ccccc|c}
            \toprule
            Variants                & Obajverse & L1   & L2   & L3   & L4   & L5   & Average \\ \hline
            \multirow{2}{*}{"-Cat"} & \XSolidBrush        & 23.7 & 25.9 & 27.4 & 28.6 & 29.0 & 26.9 \\
                                    & \Checkmark       & 20.4\textcolor{blue}{$^{\downarrow 3.3}$} & 22.3\textcolor{blue}{$^{\downarrow 3.6}$} & 22.9\textcolor{blue}{$^{\downarrow 4.5}$} & 21.6\textcolor{blue}{$^{\downarrow 7.0}$} & 23.3\textcolor{blue}{$^{\downarrow 5.7}$} & 22.1\textcolor{blue}{$^{\downarrow 4.8}$} \\ \hline
            \multirow{2}{*}{"-Sep"} & \XSolidBrush        & 17.6 & 20.1 & 18.1 & 16.4 & 15.3 & 20.3 \\
                                    & \Checkmark       & 15.3\textcolor{blue}{$^{\downarrow 2.3}$} & 17.2\textcolor{blue}{$^{\downarrow 2.9}$} & 17.1\textcolor{blue}{$^{\downarrow 1.0}$} & 14.3\textcolor{blue}{$^{\downarrow 2.1}$} & 12.8\textcolor{blue}{$^{\downarrow 2.5}$} & 15.3\textcolor{blue}{$^{\downarrow 5.0}$} \\
            \bottomrule
            \end{tabular}
    \end{subtable}
\label{tab:objaverse}
\end{table*}

\subsection{Experimental Details}
\label{sub_sec:exp_details}

\noindent \textbf{Implementation Details.} 
We sample $N=1024$ points of each point cloud and divide it into $N_{C}=64$ point patches, each with $M=32$ neighborhood points. Besides, we set the mask ratio as $0.7$. We randomly select a prompt pair that performs the same task with the query point cloud from the training set. We consider $R_{rec} \in \{ 3\%, 6\%, 12\%, 25\%, 50\% \}$ and $R_{den} \in \{ 10\%, 20\%, 30\%, 40\%, 50\% \}$. We use an AdamW optimizer~\cite{adam} and cosine learning rate decay, with the initial learning rate as 0.001 and a weight decay of 0.05. \textcolor{blue}{For PIC++, we} use a pre-trained checkpoint from PIC as model initialization. All models are trained for 300 epochs.

\noindent \textbf{Baseline Methods and Training Details}
To evaluate the proposed framework, we evaluate the two variants~(``-Sep'' and ``-Cat'') of PIC and PIC++ on the datasets mentioned in Sec.~\ref{sec:in-context_datasets}. We compare them with the following methods:

\noindent \textbf{1) Task-Specific Models.}
We select three representative methods: PointNet~\cite{pointnet}, DGCNN~\cite{dgcnn}, PCT~\cite{pct}, and ACT~\cite{act}, and train them individually on each task or dataset. For the multitasking benchmark, we also design task-specific heads for each task.

\noindent \textbf{2) Multitask Models$^{\dagger}$.}
For a fair comparison, we develop multitask models based on common point cloud analysis methods. These models feature a shared backbone network and multiple task-specific heads designed to be capable of multitasking learning.

\begin{table*}[h]
\scriptsize
\caption{\textbf{Comparison results on Multi-Entity Segmentation In-Context Datasets}, including ShapeNetPart~\cite{shapenetpart}, Human3D~\cite{human3d}, and BEHAVE~\cite{behave}. Additionally, we conduct one-shot generalization testing experiments on AKB-48~\cite{akb48}, which is not included in the training set. $\dagger$ represents training from scratch, $\ddagger$ represents full fine-tuning. We report the instance mIoU metric.}
\setlength\tabcolsep{3mm}
\renewcommand\arraystretch{1.2}
\centering
\begin{tabular}{llccccccc}
\toprule
\multicolumn{2}{l|}{\multirow{3}{*}{Models}} & \multicolumn{1}{c|}{\multirow{3}{*}{Venues}} & \multicolumn{2}{c|}{ShapeNetPart}     & \multicolumn{2}{c|}{Human3D}    & \multicolumn{1}{c|}{BEHAVE} & \cellcolor{blue!10}{AKB-48} \\
\multicolumn{2}{l|}{}            & \multicolumn{1}{c|}{}              & mIoU$\uparrow$ & \multicolumn{1}{c|}{mIoU$\uparrow$} & mIoU$\uparrow$  & \multicolumn{1}{c|}{mIoU$\uparrow$} & \multicolumn{1}{c|}{mIoU$\uparrow$}    & \cellcolor{blue!10}{mIoU$\uparrow$} \\
\multicolumn{2}{l|}{}            & \multicolumn{1}{c|}{}              & Test & \multicolumn{1}{c|}{Val} & Test  & \multicolumn{1}{c|}{Val} & \multicolumn{1}{c|}{Test}    & \cellcolor{blue!10}{}    \\ \hline
\rowcolor{gray!10}
\multicolumn{9}{c}{Single Dataset Supervised Training}        \\ \hline
\rowcolor{gray!10}
\multicolumn{2}{l|}{PointNet~\cite{pointnet}}    & \multicolumn{1}{c|}{CVPR'17}       & 77.45    & \multicolumn{1}{c|}{80.12}    & 81.05    & \multicolumn{1}{c|}{83.91}    & \multicolumn{1}{c|}{78.63}  & \cellcolor{blue!10}{/}   \\
\rowcolor{gray!10}
\multicolumn{2}{l|}{DGCNN~\cite{dgcnn}}       & \multicolumn{1}{c|}{TOG'19}        & 76.16    & \multicolumn{1}{c|}{78.23}    & 74.57    & \multicolumn{1}{c|}{76.28}    & \multicolumn{1}{c|}{86.18}  & \cellcolor{blue!10}{/}   \\
\rowcolor{gray!10}
\multicolumn{2}{l|}{PCT~\cite{pct}}         & \multicolumn{1}{c|}{CVM'21}        & 79.46    & \multicolumn{1}{c|}{83.71}    & 81.65    & \multicolumn{1}{c|}{83.16}    & \multicolumn{1}{c|}{79.42}  & \cellcolor{blue!10}{/}   \\ \hline
\multicolumn{9}{c}{Multi-Dataset Joint Training}        \\ \hline
\multicolumn{2}{l|}{PointNet~\cite{pointnet}}    & \multicolumn{1}{c|}{CVPR'17}       & 75.77    & \multicolumn{1}{c|}{79.07}    & 77.44    & \multicolumn{1}{c|}{79.33}    & \multicolumn{1}{c|}{66.06}  & \cellcolor{blue!10}{41.38}   \\
\multicolumn{2}{l|}{DGCNN~\cite{dgcnn}}       & \multicolumn{1}{c|}{TOG'19}        & 75.16    & \multicolumn{1}{c|}{78.58}    & 70.52    & \multicolumn{1}{c|}{74.55}    & \multicolumn{1}{c|}{77.48}  & \cellcolor{blue!10}{36.07}   \\
\multicolumn{2}{l|}{PCT~\cite{pct}}         & \multicolumn{1}{c|}{CVM'21}        & 79.65    & \multicolumn{1}{c|}{82.00}    & 79.46    & \multicolumn{1}{c|}{80.75}    & \multicolumn{1}{c|}{72.41}  & \cellcolor{blue!10}{40.17}   \\
\multicolumn{2}{l|}{PTv1~\cite{pointtransformer}}    & \multicolumn{1}{c|}{ICCV'21}       & 80.12     & \multicolumn{1}{c|}{83.24}    & 79.64   & \multicolumn{1}{c|}{81.11}    & \multicolumn{1}{c|}{77.46}  & \cellcolor{blue!10}{36.47}   \\
\multicolumn{2}{l|}{PointMLP~\cite{pointmlp}}    & \multicolumn{1}{c|}{ICLR'22}       & 83.76     & \multicolumn{1}{c|}{86.32}    & 81.34   & \multicolumn{1}{c|}{84.64}    & \multicolumn{1}{c|}{84.56}  & \cellcolor{blue!10}{30.86}   \\
\multicolumn{2}{l|}{PointNeXt~\cite{pointnext}}    & \multicolumn{1}{c|}{NeurIPS'22}       & 79.87     & \multicolumn{1}{c|}{82.92}    & 82.06   & \multicolumn{1}{c|}{84.52}    & \multicolumn{1}{c|}{76.97}  & \cellcolor{blue!10}{37.39}   \\
\multicolumn{2}{l|}{PTv2~\cite{pointtransformerv2}}    & \multicolumn{1}{c|}{NeurIPS'22}       & 81.74     & \multicolumn{1}{c|}{83.18}    & 80.84   & \multicolumn{1}{c|}{82.15}    & \multicolumn{1}{c|}{82.63}  & \cellcolor{blue!10}{33.67}   \\
\multicolumn{2}{l|}{PointM2AE~\cite{pointm2ae}}    & \multicolumn{1}{c|}{NeurIPS'22}       & 77.37     & \multicolumn{1}{c|}{79.46}    & 76.20   & \multicolumn{1}{c|}{78.36}    & \multicolumn{1}{c|}{78.31}  & \cellcolor{blue!10}{39.30}   \\
\multicolumn{2}{l|}{ACT~\cite{act}}    & \multicolumn{1}{c|}{ICLR'23}       & 84.43    & \multicolumn{1}{c|}{85.98}    & 82.58    & \multicolumn{1}{c|}{84.69}    & \multicolumn{1}{c|}{88.33}  & \cellcolor{blue!10}{34.69}   \\
\multicolumn{2}{l|}{I2P-MAE~\cite{i2pmae}}    & \multicolumn{1}{c|}{CVPR'23}       & 74.84    & \multicolumn{1}{c|}{77.68}    & 74.36    & \multicolumn{1}{c|}{78.57}    & \multicolumn{1}{c|}{79.00}  & \cellcolor{blue!10}{16.89}   \\
\multicolumn{2}{l|}{MAMBA3D~\cite{han2024mamba3d}}    & \multicolumn{1}{c|}{MM'24}       & 82.26    & \multicolumn{1}{c|}{83.51}    &  78.92   & \multicolumn{1}{c|}{82.36}    & \multicolumn{1}{c|}{82.27}  & \cellcolor{blue!10}{42.25}   \\
\multicolumn{2}{l|}{PointDiff~\cite{pointdiff}}    & \multicolumn{1}{c|}{CVPR'24}       & 80.64     & \multicolumn{1}{c|}{82.14}    & 80.32   & \multicolumn{1}{c|}{83.04}    & \multicolumn{1}{c|}{80.88}  & \cellcolor{blue!10}{37.36}   \\
\multicolumn{2}{l|}{PTv3~\cite{pointtransformerv3}}    & \multicolumn{1}{c|}{CVPR'24}       & 82.37     & \multicolumn{1}{c|}{82.45}    & 81.04   & \multicolumn{1}{c|}{83.13}    & \multicolumn{1}{c|}{85.63}  & \cellcolor{blue!10}{44.42}   \\
\multicolumn{2}{l|}{PointLoRA~\cite{pointlora}}    & \multicolumn{1}{c|}{CVPR'25}       & 76.17     & \multicolumn{1}{c|}{78.47}    & 76.17   & \multicolumn{1}{c|}{79.02}    & \multicolumn{1}{c|}{76.16}  & \cellcolor{blue!10}{37.27}   \\
\multicolumn{2}{l|}{PCMamba~\cite{pcm}}    & \multicolumn{1}{c|}{AAAI'25}       & 81.39    & \multicolumn{1}{c|}{83.89}    & 81.22  & \multicolumn{1}{c|}{83.74}    & \multicolumn{1}{c|}{83.64}  & \cellcolor{blue!10}{38.18}   \\
\hline
\multicolumn{2}{l|}{Copy}    & \multicolumn{1}{c|}{/}       & 23.75    & \multicolumn{1}{c|}{23.18}    & 8.94    & \multicolumn{1}{c|}{9.12}    & \multicolumn{1}{c|}{37.27}  & \cellcolor{blue!10}{17.24}   \\ \hline
\multirow{2}{*}{"-Cat"} & \multicolumn{1}{l|}{PIC}     & \multicolumn{1}{c|}{NeurIPS'23}              & 64.52    & \multicolumn{1}{c|}{70.50}    & 72.79    & \multicolumn{1}{c|}{75.28}    & \multicolumn{1}{c|}{83.66}  &  \cellcolor{blue!10}{9.97}  \\
& \multicolumn{1}{l|}{PIC++$^\dagger$}   & \multicolumn{1}{c|}{/}              & 81.13\textbf{\textcolor{blue}{$^{\uparrow}$}}   & \multicolumn{1}{c|}{85.32\textbf{\textcolor{blue}{$^{\uparrow}$}}}    & 77.33\textbf{\textcolor{blue}{$^{\uparrow}$}}    & \multicolumn{1}{c|}{79.13\textbf{\textcolor{blue}{$^{\uparrow}$}}}    & \multicolumn{1}{c|}{86.22\textbf{\textcolor{blue}{$^{\uparrow}$}}}  & \cellcolor{blue!10}{73.11\textbf{\textcolor{blue}{$^{\uparrow}$}}}   \\
& \multicolumn{1}{l|}{PIC++$^\ddagger$}   & \multicolumn{1}{c|}{/}              & 84.13\textbf{\textcolor{blue}{$^{\uparrow}$}}    & \multicolumn{1}{c|}{87.31\textbf{\textcolor{blue}{$^{\uparrow}$}}}    & 80.11\textbf{\textcolor{blue}{$^{\uparrow}$}}    & \multicolumn{1}{c|}{83.34\textbf{\textcolor{blue}{$^{\uparrow}$}}}    & \multicolumn{1}{c|}{88.03\textbf{\textcolor{blue}{$^{\uparrow}$}}}  &  \cellcolor{blue!10}{\textbf{73.52\textbf{\textcolor{blue}{$^{\uparrow}$}}}}  \\ \hline
\multirow{2}{*}{"-Sep"} & \multicolumn{1}{l|}{PIC}     & \multicolumn{1}{c|}{NeurIPS'23}              & 79.22    & \multicolumn{1}{c|}{81.72}    & 74.74    & \multicolumn{1}{c|}{77.93}    & \multicolumn{1}{c|}{84.03}  & \cellcolor{blue!10}{19.58}   \\
& \multicolumn{1}{l|}{PIC++$^\dagger$}   & \multicolumn{1}{c|}{/}              & 80.64\textbf{\textcolor{blue}{$^{\uparrow}$}}   & \multicolumn{1}{c|}{83.26\textbf{\textcolor{blue}{$^{\uparrow}$}}}    & 80.39\textbf{\textcolor{blue}{$^{\uparrow}$}}    & \multicolumn{1}{c|}{82.36\textbf{\textcolor{blue}{$^{\uparrow}$}}}    & \multicolumn{1}{c|}{87.74\textbf{\textcolor{blue}{$^{\uparrow}$}}}  & \cellcolor{blue!10}{68.96\textbf{\textcolor{blue}{$^{\uparrow}$}}}   \\
& \multicolumn{1}{l|}{PIC++$^\ddagger$}   & \multicolumn{1}{c|}{/}              & \textbf{85.14\textbf{\textcolor{blue}{$^{\uparrow}$}}}    & \multicolumn{1}{c|}{\textbf{87.82\textbf{\textcolor{blue}{$^{\uparrow}$}}}}    & \textbf{82.82\textbf{\textcolor{blue}{$^{\uparrow}$}}}    & \multicolumn{1}{c|}{\textbf{85.59\textbf{\textcolor{blue}{$^{\uparrow}$}}}}    & \multicolumn{1}{c|}{\textbf{88.63\textbf{\textcolor{blue}{$^{\uparrow}$}}}}  & \cellcolor{blue!10}{63.08\textbf{\textcolor{blue}{$^{\uparrow}$}}}   \\ 
\bottomrule
\end{tabular}
\label{tab:main_result_2}
\vspace{-1em}
\end{table*}

\noindent \textbf{3) Multitask Models$^{\ddagger}$.} For \textbf{PointNet}~\cite{pointnet}, \textbf{DGCNN}~\cite{dgcnn}, and \textbf{PCT}~\cite{pct}, we utilize a pre-trained backbone that is trained on the ShapeNet [7] dataset for classification tasks. This pre-trained backbone network is equipped with multiple task-specific heads to perform multitask learning on our benchmark. \textbf{PointMAE}~\cite{pointmae} is a masked auto-encoder. Unlike Point-BERT, PointMAE directly rebuilds points in each local area. \textbf{ACT}~\cite{act}, \textbf{I2P-MAE}~\cite{i2pmae}, and \textbf{ReCon}~\cite{recon} are recent SoTA methods that involve other modalities like image and text knowledge in the pre-training stage and enhance the performance on different tasks after fine-tuning the models. Similar to multitask models, we combine a pre-trained encoder with different task heads for training on the four tasks.

\noindent \textbf{Copy Example} is a baseline that uses the prompt's target point cloud as its prediction.

\subsection{Results on Point Cloud Multitasking}
\label{sub_sec:result_on_multitasking}

\begin{figure*}[t]
\hsize=\textwidth
\centering
\includegraphics[width=0.99\textwidth]{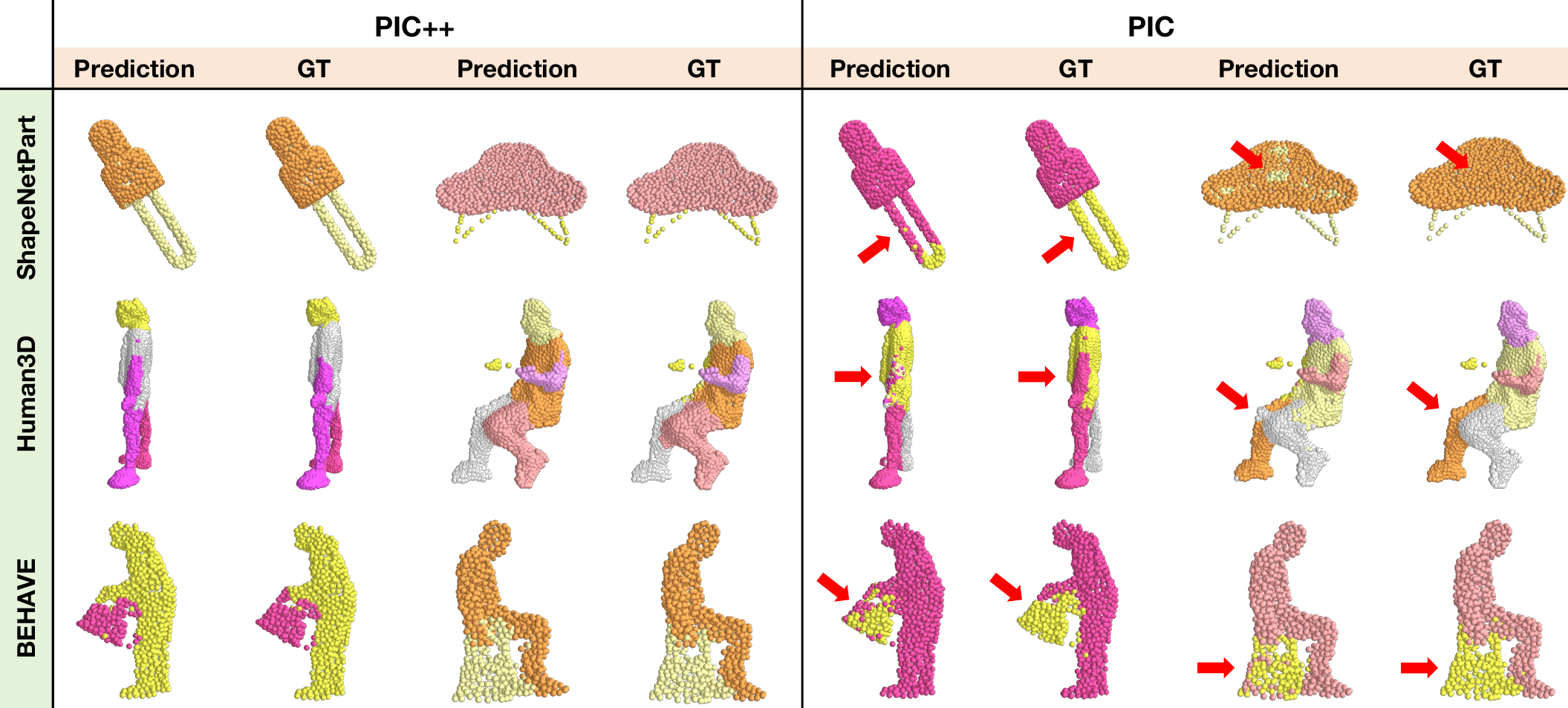}
\caption{
\textbf{Visualization of comparison results between PIC++(extended version) and PIC~(vanilla version).} PIC++ generates more accurate segmentation results than PIC, as can be seen from the spots with red arrows (\bm{\textcolor{red}{$\searrow$}}) pointed to them.
}
\label{fig:comparison_vis_seg}
\vspace{-1em}
\end{figure*}

We report extensive experimental results of various models on the dataset we proposed in Tab.~\ref{tab:main_result_1}. From this, we find that our PICs exhibits impressive results and are capable of adapting to different tasks after only one training, achieving SoTA results in all four tasks.
Besides, we visualize the in-context 3D inference results of the "-Sep" variant of PIC in Fig.~\ref{fig:visualization_main}, where our model can generate the corresponding predictions given the provided prompts among all tasks, including reconstruction, denoising, registration, and part segmentation. 

\noindent \textbf{Comparison to Task-Specific Models.}
Task-specific models, including PointNet~\cite{pointnet}, DGCNN~\cite{dgcnn}, PCT~\cite{pct}, and ACT~\cite{act} outperform PIC in most indicators on all tasks. However, in the part segmentation task, PIC++ achieves better results than PointNet, DGCNN, PCT, and ACT, though they are trained for segmentation. It needs to be clarified that the performance of our model is directly related to the choice of prompts. When the quality of the prompt is better, PIC can achieve much better results, and this will be demonstrated in the following section.

\begin{figure*}[t]
\centering
\includegraphics[width=0.8\textwidth]{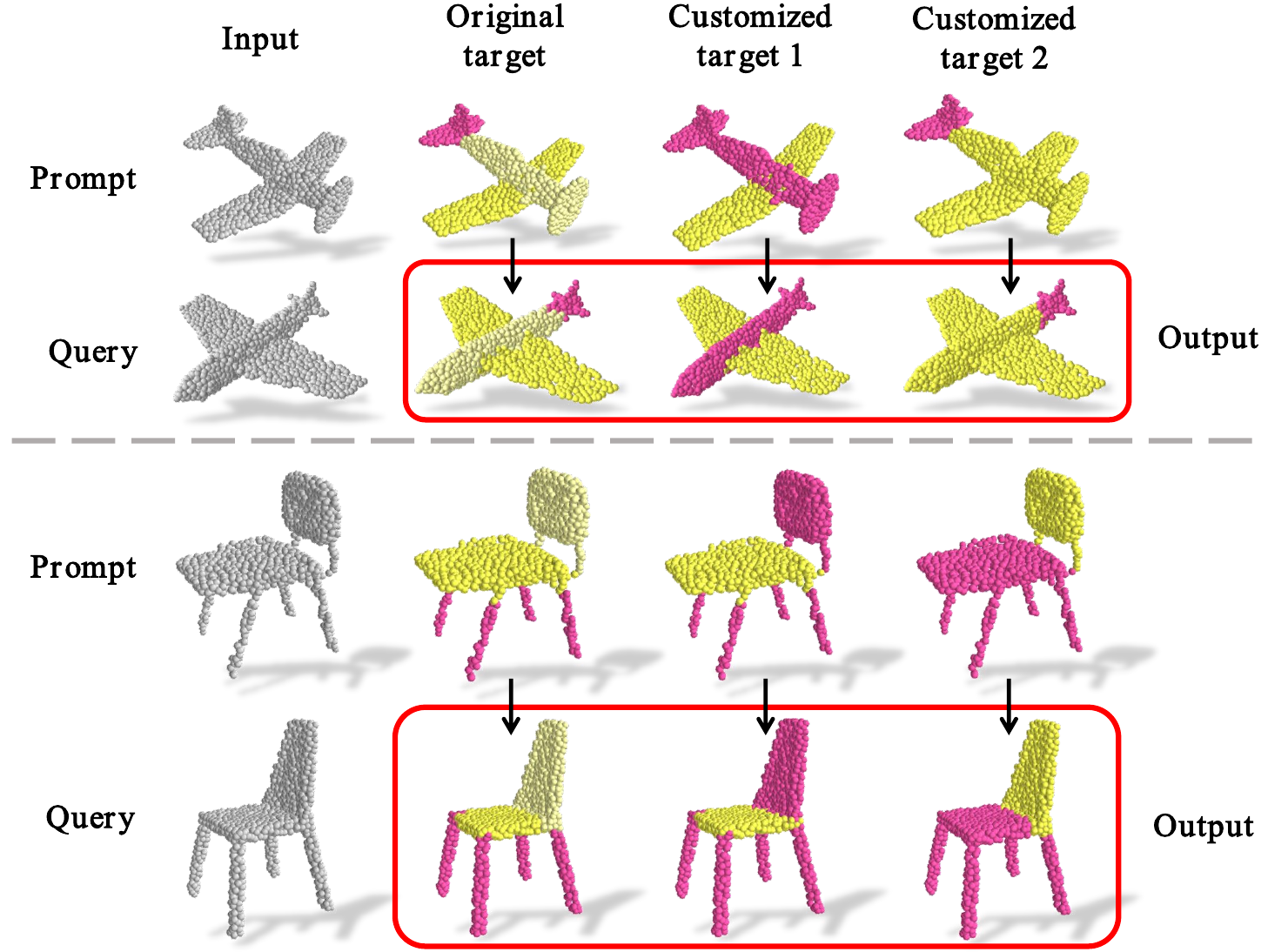}
\caption{
\textbf{Generalization of PIC++.} We use customized prompts to guide the model to perform specified part segmentation. The red boxes indicate the output of the "-Sep" variant of PIC++.
}
\label{fig:generalization_seg}
\vspace{-1em}
\end{figure*}

\noindent \textbf{Comparison to Multitask Models.}
As shown in Tab.~\ref{tab:main_result_1}, our PIC and PIC++ outperform multitask models in multitasking learning, especially in part segmentation, while the multitask models suffer from confusion from multiple types of tasks during training. 
We also compare visualization results between our PIC models and multitask models on three tasks, including reconstruction, denoising, and registration. 
As shown in Fig.~\ref{fig:supp_comparison}, our PIC-Sep and PIC-Cat outputs are more satisfactory than other multitask models.

\noindent \textbf{Generalization.}
Usually, in-context learning has a certain degree of generalization ability, thus allowing the model to adapt to different tasks quickly. This also applies to our models.
First, we test the proposed methods on out-of-distribution point clouds by conducting a registration evaluation on the ModelNet40~\cite{modelnet40} dataset.
As shown in Fig.~\ref{fig:transfer_ability}~(a), our models, PIC-Sep and PIC-Cat, both exhibit superior performance on open-class tasks compared to supervised learning models specifically trained for the registration task. We compare it with models trained on the single task, such as PointMAE~\cite{pointmae}, and PCT~\cite{pct}. These models suffer obvious performance drops when transferred to the new dataset. The greater the difficulty, the greater the decline in performance.

Furthermore, we validate their generalization abilities on unseen tasks, such as any angle registration and local completion. As shown in Fig.~\ref{fig:transfer_ability}~(b), our proposed models work well on both tasks, validating their abilities to transfer learned knowledge. As a comparison, task-specific models are unable to infer any angle-rotated point cloud, even if they are trained on the registration. 

\noindent \textbf{Training on Large-Scale Objaverse~\cite{objaverse}.}
To assess the impact of data scale on the model, we incorporate a large-scale dataset, Objaverse, into the training set. This dataset comprises 800K point clouds, with each sample including only XYZ coordinates. During training, we construct input-target pairs for Objaverse point clouds in the forms of point cloud reconstruction, denoising, and registration, and perform joint training with the original training data. As shown in Tab.~\ref{tab:objaverse} (a), the performance of PIC improves significantly after incorporating Objaverse. Additionally, we test the generalization of the above model on ModelNet40. As shown in Tab.~\ref{tab:objaverse} (b), the ability of PIC to generalize to unseen datasets improves considerably after adding the large-scale data. Overall, these results indicate that increasing the data scale helps to enhance the model's in-context learning ability.

\begin{table*}[t]
    \centering
    \renewcommand\arraystretch{1.1}
    \caption{\textbf{Effectiveness of Joint Sampling Module.} We conduct experiments on ShapeNet In-Context Datasets using PIC. We report the results of denoising and part segmentation.}
    \setlength\tabcolsep{3mm}
    \begin{tabular}{l|c|cc}
        \toprule
        Variants   & Joint Sampling      & Denoising CD$\downarrow$  & Part Segmentation mIoU$\uparrow$ \\  \hline
        \multirow{2}{*}{"-Cat"} & \XSolidBrush    & 29.3         & 17.03                        \\
                                 &  \Checkmark     & \textbf{5.3}       &  \textbf{78.95}                     \\ \hline
        \multirow{2}{*}{"-Sep"} & \XSolidBrush   &  36.3       &  23.72                       \\
                                 &  \Checkmark    &  \textbf{7.6}        &  \textbf{74.95}                     \\ \bottomrule
    \end{tabular}
    \label{tab:JSmodule}
\end{table*}

\subsection{Results on Multi-Entity Segmentation}
\label{sub_sec:result_on_segmentation}

\noindent \textbf{Comparison to Other Methods.}
We compare our PIC and PIC++ with other methods, all of which have well-designed model architectures and segmentation task heads. As shown in Tab.~\ref{tab:main_result_2}, the part segmentation performance of PIC under multi-dataset joint training is suboptimal. To ensure a fair comparison, we additionally conduct experiments on PIC++ trained from scratch (denoted by $\dagger$). As illustrated in the table, both versions of PIC++ achieve competitive results. Specifically, the scratch-trained PIC++ consistently outperforms the baseline PIC, while the fine-tuned PIC++ (denoted by $\ddagger$) surpasses all other models and achieves SoTA results on Multi-Entity In-Context Datasets. Whether in in-domain or out-of-domain datasets, this result indicates that our extended version, PIC++, is more adept at integrating multiple datasets in segmentation tasks.

\noindent \textbf{One-Shot Testing on AKB-48.}
In addition to the part segmentation datasets included in the training set (ShapeNetPart, Human3D, BEHAVE), we also conduct one-shot generalization tests on AKB-48 to assess the model's ability to generalize to out-of-domain datasets. For our PIC models, we randomly select one sample from each category as our prompt, ensuring that the prompt has the same parts as the query point cloud. For other models, we randomly select one sample from each category and use the corresponding pre-trained model to extract features for each part, which serve as the segmentation standard. Later, the predicted results are classified for each point by calculating the cosine similarity between the predicted and extracted features.

As shown in the last column of Tab.\ref{tab:main_result_2}, our PIC++ not only achieves SoTA results on the in-domain datasets(ShapeNetPart, Human3D, BEHAVE), but also exhibits strong generalization capabilities on out-of-domain datasets~(AKB-48). Specifically, PIC++ trained from scratch achieves impressive generalization performance (73.11\% for -Cat'' and 68.96\% for-Sep''), which is comparable to the fine-tuned version's 73.52\% (-Cat'') and even superior to its 63.08\% (-Sep''). Overall, PIC++ surpasses PIC and other models by a large margin, showing superior generalization performance.

\begin{table*}[t]
    \caption{\textbf{Influence of prompt selecting strategies.} We explore the influence of prompt quality on the model's performance. We conduct comparative experiments using PIC and PIC++ on the ShapeNet In-Context Datasets and Multi-Entity In-Context Datasets, respectively. Note that AKB-48 is not included in the training set. Gray indicates our default settings in Tab.~\ref{tab:main_result_1} and Tab.~\ref{tab:main_result_2}.}
    \centering
    
    \begin{subtable}[t]{0.99\linewidth}
        \renewcommand\arraystretch{1.1}
        \setlength\tabcolsep{1.9mm}
        \caption{Experimental results of PIC on ShapeNet In-Context Datasets.}
        \centering
        \begin{tabular}{c|c|ccc}
            \toprule
            Variants   & Selection method & Reconstruction CD$\downarrow$ & Denoising CD$\downarrow$ & Registration CD$\downarrow$ \\ \hline
            \multirow{4}{*}{"-Cat"} & \cellcolor{gray!10} Random     & \cellcolor{gray!10} 4.3    & \cellcolor{gray!10} 5.3   & \cellcolor{gray!10} 14.1     \\
                                     & Class-aware  & 4.3 & 5.3  & 10.5  \\
                                     & Fea-aware  & 4.3 & 5.3  & 10.7  \\
                                     & CD-aware  & \textbf{4.3} & \textbf{5.3}  & \textbf{9.6}  \\ \hline
            \multirow{4}{*}{"-Sep"} & \cellcolor{gray!10} Random      & \cellcolor{gray!10} 4.7  & \cellcolor{gray!10} 7.6  & \cellcolor{gray!10} 10.3 \\
                                     & Class-aware & 4.6 & 7.4  & 5.1 \\ 
                                     & Fea-aware & 4.9 & 7.6  & 5.8 \\ 
                                     & CD-aware & \textbf{4.4} & \textbf{7.1}  & \textbf{4.1} \\ \bottomrule
            \end{tabular}
    \end{subtable}
    
    \begin{subtable}[t]{0.99\linewidth}
        \renewcommand\arraystretch{1.1}
        \setlength\tabcolsep{3mm}
        \caption{Experimental results of PIC++ on Multi-Entity In-Context Datasets.}
        \centering
        \begin{tabular}{c|c|cccc}
            \toprule
            Variants   & Selection method & \begin{tabular}[c]{@{}c@{}}ShapeNetPart\\ mIoU$\uparrow$\end{tabular} & \begin{tabular}[c]{@{}c@{}}Human3D\\ mIoU$\uparrow$\end{tabular} & \begin{tabular}[c]{@{}c@{}}BEHAVE\\ mIoU$\uparrow$\end{tabular} & \begin{tabular}[c]{@{}c@{}} \cellcolor{blue!10} AKB-48\\  \cellcolor{blue!10} mIoU$\uparrow$\end{tabular} \\ \hline
            \multirow{3}{*}{"-Cat"} & \cellcolor{gray!10} Random     & \cellcolor{gray!10} 84.13    & \cellcolor{gray!10} 80.11   & \cellcolor{gray!10} 88.03  & \cellcolor{blue!10} 73.52  \\
                                     & Fea-aware  & 85.51 & 80.91  & 88.04  & \cellcolor{blue!10} 75.37\\
                                     & CD-aware  & \textbf{88.14} & \textbf{82.11}  & \textbf{88.21} & \cellcolor{blue!10} \textbf{79.27} \\ \hline
            \multirow{3}{*}{"-Sep"} & \cellcolor{gray!10} Random      & \cellcolor{gray!10} 85.14  & \cellcolor{gray!10} 82.82  & \cellcolor{gray!10} 88.63  & \cellcolor{blue!10} 63.08 \\
                                     & Fea-aware & 86.10 & 82.91  & 88.64 & \cellcolor{blue!10} 70.26 \\ 
                                     & CD-aware & \textbf{87.78} & \textbf{83.89}  & \textbf{88.72}  & \cellcolor{blue!10} \textbf{77.11} \\ \bottomrule
            \end{tabular}
    \end{subtable}
\label{tab:prompt_selecting}
\vspace{-1em}
\end{table*}

\begin{table*}[t]
    \centering
    \setlength\tabcolsep{1.8mm}

    \caption{\textbf{Ablation study on mask ratio.} We conduct ablation experiments on the separated version of PIC~("-Sep") on the ShapeNet In-Context Datasets. For reconstruction, denoising, and registration, we report the CD loss (x1000). For part segmentation, we report the mIoU metric.}
    \begin{tabular}{c|c|cccc}
        \toprule
        \# & Mask Ratio & Reconstruction CD$\downarrow$ & Denoising CD$\downarrow$ & Registration CD$\downarrow$ & Part Seg. mIoU$\uparrow$ \\ \hline
        1  & 0.2     & 27.8  & 33.5     & 68.2     & 43.84    \\
        2  & 0.3     & 5.2  & \textbf{7.3}     & 14.8     & 56.72    \\
        3  & 0.4     & 5.0  & 7.4     & 12.3     & 60.25    \\
        4  & 0.5     & \textbf{4.0}  & 7.5     & 11.5     & 64.68    \\
        5  & 0.6     & 4.9         & 7.8       & \textbf{9.4}     & 70.17   \\
        \rowcolor{gray!10}
        6  & 0.7     & 4.7         & 7.6        & 10.3     & \textbf{74.95}     \\ \bottomrule
    \end{tabular}

    \label{tab:mask_ratio}
\vspace{-1em}
\end{table*}

\begin{table*}[t]
    \small
    \caption{\textbf{Ablation study on PIC.} We conduct ablation experiments on ShapeNet In-Context Datasets. Gray: default setting.}
    \vspace{-0.5em}
    \centering
    \begin{minipage}{0.32\linewidth}
        \renewcommand\arraystretch{1.1}
        \centering
        \subfloat[Different sampling strategies.]{
        \resizebox{1\textwidth}{!}{%
        \begin{tabular}{c|c|cccc}
            \toprule
            \multicolumn{1}{l|}{\#} & \multicolumn{1}{l|}{\begin{tabular}[c]{@{}l@{}}Sample\\ method\end{tabular}} & \begin{tabular}[c]{@{}c@{}}Rec.\\ CD$\downarrow$\end{tabular} & \begin{tabular}[c]{@{}c@{}}Den.\\ CD$\downarrow$\end{tabular} & \begin{tabular}[c]{@{}c@{}}Reg.\\ CD$\downarrow$\end{tabular} & \begin{tabular}[c]{@{}c@{}}Part Seg.\\ mIoU$\uparrow$\end{tabular} \\ \hline
            1     & RS      & 42.6      & 11.6       & 12.1    & \textbf{77.24}       \\
            \rowcolor{gray!10}
            2     & FPS  & \textbf{4.7}  & \textbf{7.6}  & \textbf{10.3} & 74.95  \\ \bottomrule
        \end{tabular}
        }}
    \end{minipage}
    \hspace{1mm}
    \begin{minipage}{0.32\linewidth}
        \renewcommand\arraystretch{1.1}
        \centering
        \subfloat[Prompt position.]{
        \resizebox{1\textwidth}{!}{%
        \begin{tabular}{c|c|cccc}
            \toprule
            \# & order  & \begin{tabular}[c]{@{}c@{}}Rec.\\ CD$\downarrow$\end{tabular} & \begin{tabular}[c]{@{}c@{}}Den.\\ CD$\downarrow$\end{tabular} & \begin{tabular}[c]{@{}c@{}}Reg.\\ CD$\downarrow$\end{tabular} & \begin{tabular}[c]{@{}c@{}}Part Seg.\\ mIoU$\uparrow$\end{tabular} \\ \hline
            1 & behind & 4.8           & 8.2          & \textbf{8.0}        & 74.04              \\
            \rowcolor{gray!10}
            2 & before & \textbf{4.7}           & \textbf{7.6}          & 10.3       & \textbf{74.95}          \\ \bottomrule
        \end{tabular}
        }}
    \end{minipage}
    \hspace{1mm}
    \begin{minipage}{0.32\linewidth}
        \renewcommand\arraystretch{0.9}
        \centering
        \subfloat[Loss functions.]{
        \resizebox{1\textwidth}{!}{%
        \begin{tabular}{c|c|cccc}
            \toprule
            \multicolumn{1}{l|}{\#} & \multicolumn{1}{l|}{\begin{tabular}[c]{@{}c@{}}Loss\\ function\end{tabular}} & \begin{tabular}[c]{@{}c@{}}Rec.\\ CD$\downarrow$\end{tabular} & \begin{tabular}[c]{@{}c@{}}Den.\\ CD$\downarrow$\end{tabular} & \begin{tabular}[c]{@{}c@{}}Reg.\\ CD$\downarrow$\end{tabular} & \begin{tabular}[c]{@{}c@{}}Part Seg.\\ mIoU$\uparrow$\end{tabular} \\ \hline
            1     & $\ell_{1}$    &  5.0    &  8.1    & 11.1    &  72.35     \\
            \rowcolor{gray!10}
            2     & $\ell_{2}$    & \textbf{4.7}  & \textbf{7.6}  & \textbf{10.3}  & \textbf{74.95}      \\ 
            3     & $\ell_{1} + \ell_{2}$   &  5.3    & 7.9    &  13.3   &   70.46  \\ \bottomrule
        \end{tabular}
        }}
    \end{minipage}
    \label{tab:ablation_pic}
\end{table*}

\noindent \textbf{Comparison to PIC.}
Although PIC performs well on multitasking benchmarks, it falls slightly short on segmentation tasks. Firstly, due to the increasing number of cumbersome label points with the growing categories of point cloud parts, it struggles to handle multi-dataset joint training effectively. Secondly, under the circumstance of fixed label points, it fails to learn important mapping relationships from prompts during testing on out-of-domain datasets. In contrast to PIC, our proposed extension, PIC++, handles multi-dataset joint training well and demonstrates robust performance. The fact that the scratch-trained PIC++ significantly outperforms the baseline PIC on all metrics (e.g., improving AKB-48 mIoU from 19.58\% to 68.96\% in the ``-Sep'' setting) strongly validates the effectiveness of our In-Context Labeling (ICL) and In-Context Enhancing (ICE) strategies. Additionally, with the training strategy of randomly assigning labels, PIC++ is not confined to pre-defined fixed label points but learns the mapping relationship within the real-time prompts. This enables PIC++ to generalize to out-of-domain datasets and achieve excellent performance.

We also visualize the predictions of the "-Sep" model of PIC++ and PIC on the Multi-Entity In-Context Datasets, along with their corresponding target point clouds. To facilitate better observation, we assign colors to the model's output results to distinguish different parts. As shown in Fig.~\ref{fig:comparison_vis_seg}, PIC++ segments the point clouds more accurately than PIC. Since the label-point assignment methods are different between PIC++ and PIC, for convenience, we only ensure that the same parts between the output point cloud and the target point cloud are visualized using the same color.

\noindent \textbf{More Generalization Capability.}
We further explore the generalization capability of PIC++. Compared to the fixed label points of PIC, PIC++ appears to be more flexible in segmentation tasks. We customize prompts to specify part segmentation for query point clouds. As shown in Fig.~\ref{fig:generalization_seg}, we combine the parts of target point clouds in the prompt into unseen part division categories and utilize this customized prompt to guide the model in performing the same part segmentation. Our PIC++ can accurately generate unique part segmentation results via customized prompts.

\begin{figure}[ht]
\centering
\includegraphics[width=0.49\textwidth]{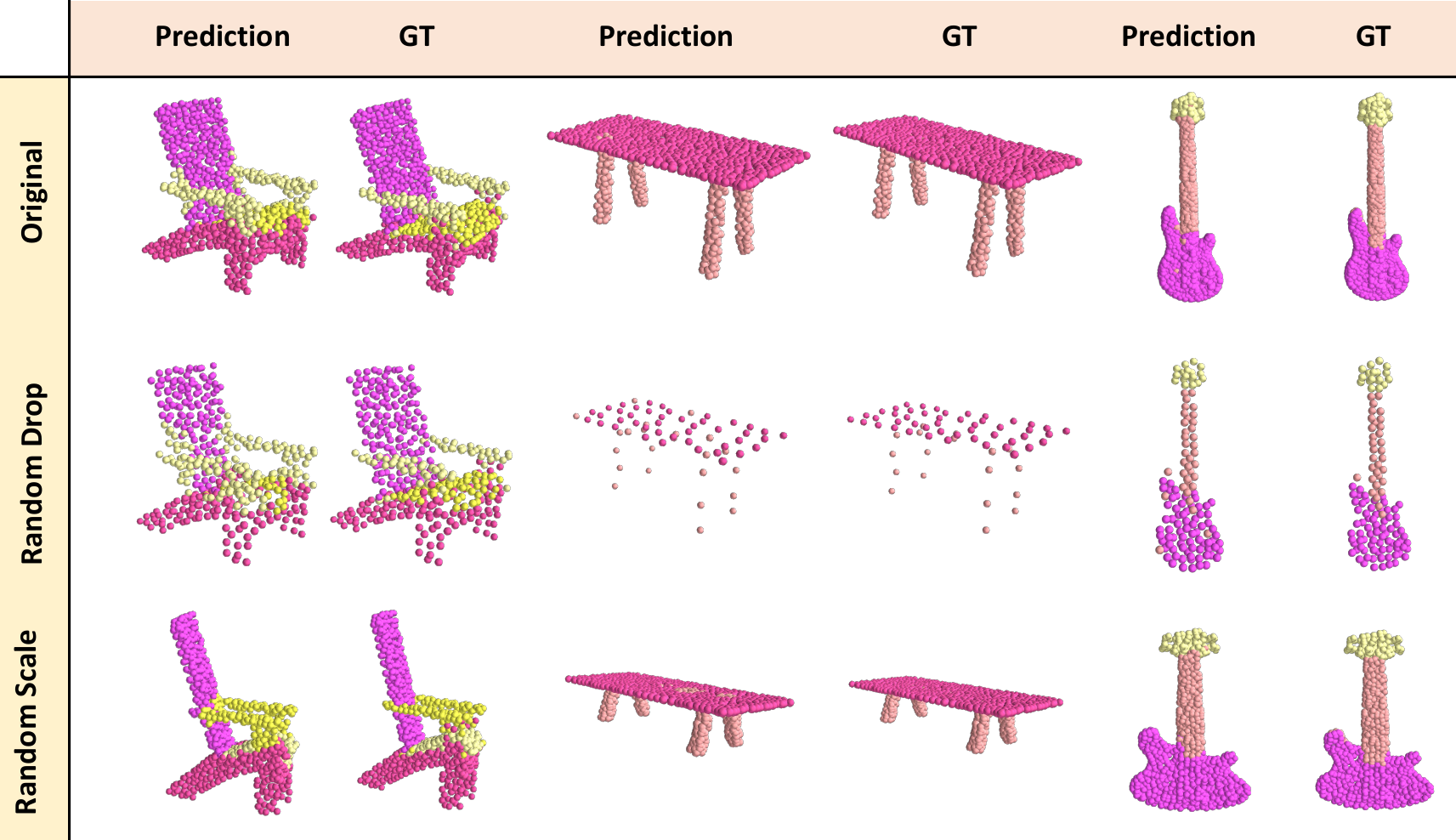}
\caption{
\textbf{Robustness on different resolutions and scales.}
}
\label{fig:robustness_inference_1}
\vspace{-1.5em}
\end{figure}

\begin{figure}[ht]
\centering
\includegraphics[width=0.49\textwidth]{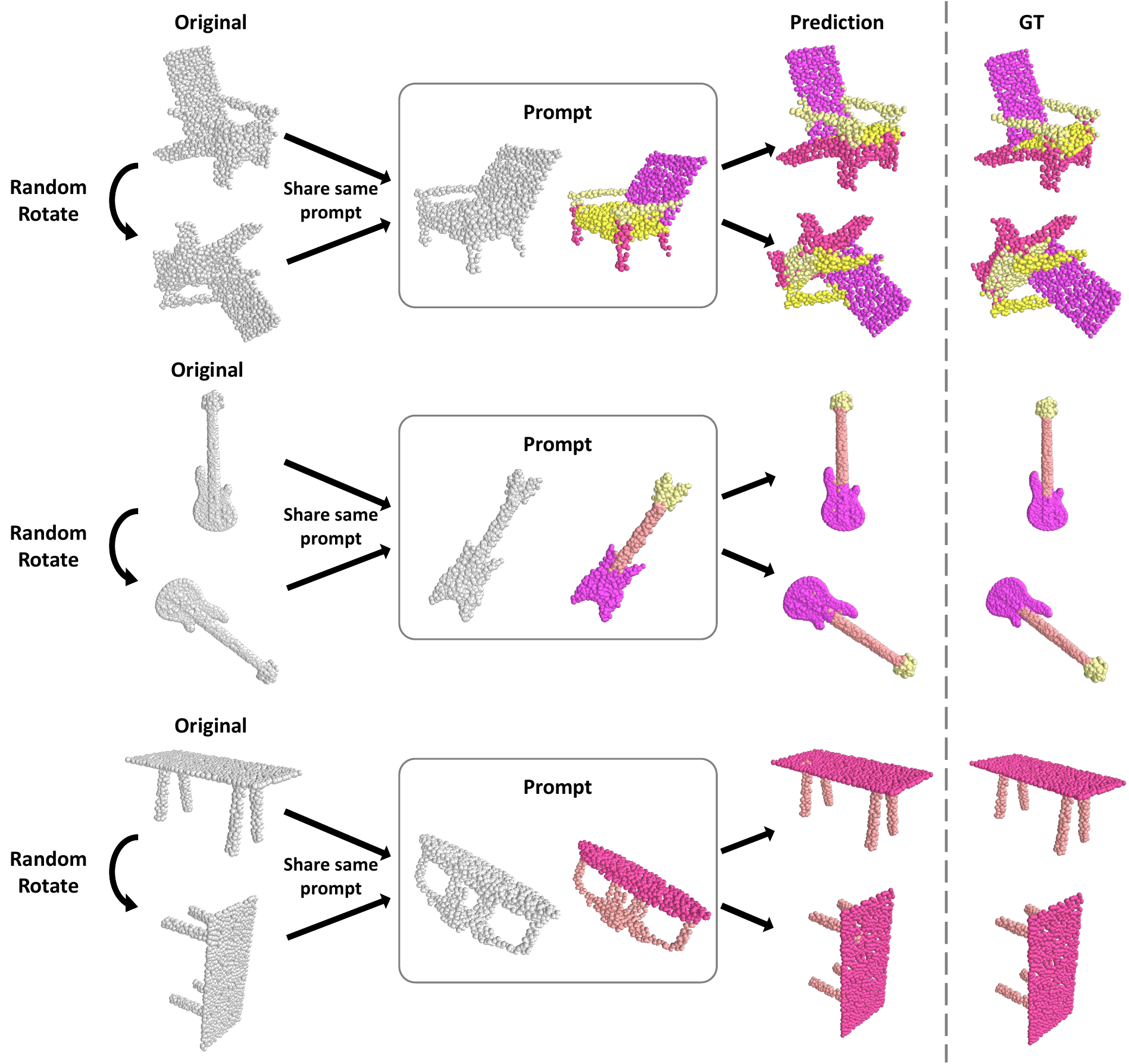}
\caption{
\textbf{Robustness on rotation.}
}
\label{fig:robustness_inference_2}
\vspace{-1.5em}
\end{figure}

\noindent \textbf{Robustness of PIC++.}
We further explored the robustness of PIC++ under various perturbations. Specifically, we subjected the query point clouds to random point dropping (simulating varying resolutions) and random scale transformations, while using the same pair of prompt point clouds to guide the part segmentation task. As illustrated in Fig.~\ref{fig:robustness_inference_1}, our model maintains excellent inference performance despite these perturbations, demonstrating the strong robustness of PIC++. Furthermore, we applied random rotations to the query point clouds while keeping the prompt conditions unchanged. As shown in Fig.~\ref{fig:robustness_inference_2}, PIC++ retains consistently high segmentation performance. These results indicate that our PIC++ does not rely on spatial alignment between the prompt and query information, but rather effectively leverages the semantic alignment of parts.

\subsection{Ablation Study}
\label{sub_sec:ablation}

\begin{table*}[t]
    \caption{\textbf{Ablation study on PIC++.} We conduct extensive experiments on the Multi-Entity In-Context Datasets using our extended version, PIC++. \textbf{(a)} shows the mIoU gap between using random prompts and ideal prompts to demonstrate the impact of In-Context Labeling~(ICL) on the upper limit of PIC++. Note that AKB-48 is still not included in the training set. \textbf{(b)} shows the effectiveness of In-Context Labeling~(ICL) and In-Context Enhancing~(ICE). \textbf{(c)} exhibits the comparison results in various choices of the loss function. Gray: default setting.}
    \vspace{-0.5em}
    \centering
    \begin{subtable}[t]{0.99\linewidth}
        \renewcommand\arraystretch{1.2}
        \caption{Influence of In-Context Labeling~(ICL) on PIC++}
        \setlength\tabcolsep{1.5mm}
        \centering
        \begin{tabular}{l|c|cccccccc}
\toprule
\multirow{2}{*}{Models}              & \multirow{2}{*}{Prompt} & \multicolumn{2}{c}{ShapeNetPart}                   & \multicolumn{2}{c}{Human3D}                        & \multicolumn{2}{c|}{BEHAVE}                         & \multicolumn{2}{c}{  AKB-48}     \\
                                     &                         & mIoU  & \multicolumn{1}{c|}{$\Delta $}                                          & mIoU  & \multicolumn{1}{c|}{$\Delta $}                                          & mIoU  & \multicolumn{1}{c|}{$\Delta $}                                          &   mIoU  &   $\Delta $                      \\ \hline
\multirow{2}{*}{\begin{tabular}[c]{@{}l@{}}"-Cat"\\ ($w/o$ ICL)\end{tabular}} & Random                  & 75.15 & \multicolumn{1}{c|}{\multirow{2}{*}{$\uparrow$ 2.86}} & 82.03 & \multicolumn{1}{c|}{\multirow{2}{*}{$\uparrow$ 1.74}} & 86.48 & \multicolumn{1}{c|}{\multirow{2}{*}{$\uparrow$ 0.05}} &   10.30 & \multirow{2}{*}{$\uparrow$ 0.39}  \\
                                     & Ideal                   & 78.01 & \multicolumn{1}{c|}{}                      & 83.77 & \multicolumn{1}{c|}{}                      & 86.53 & \multicolumn{1}{c|}{}                      &   10.69 &             \\ \hline
\multirow{2}{*}{\begin{tabular}[c]{@{}l@{}}"-Cat"\\ ($w/$ ICL)\end{tabular}}  & Random                  & 84.13 & \multicolumn{1}{c|}{\multirow{2}{*}{\textbf{$\uparrow$ 4.33}}} & 80.11 & \multicolumn{1}{c|}{\multirow{2}{*}{\textbf{$\uparrow$ 3.04}}} & 88.03 & \multicolumn{1}{c|}{\multirow{2}{*}{\textbf{$\uparrow$ 0.34}}} &   73.52 & \multirow{2}{*}{\textbf{$\uparrow$ 8.23}}  \\
                                     & Ideal                   & 88.46 & \multicolumn{1}{c|}{}                      & 83.15 & \multicolumn{1}{c|}{}                      & 88.37 & \multicolumn{1}{c|}{}                      &   81.75 &                        \\ \hline
\multirow{2}{*}{\begin{tabular}[c]{@{}l@{}}"-Sep"\\ ($w/o$ ICL)\end{tabular}} & Random                  & 80.16 & \multicolumn{1}{c|}{\multirow{2}{*}{$\uparrow$ 0.85}} & 80.85 & \multicolumn{1}{c|}{\multirow{2}{*}{$\downarrow$ 0.03}} & 84.62 & \multicolumn{1}{c|}{\multirow{2}{*}{$\downarrow$ 0.05}} &   11.77 & \multirow{2}{*}{$\uparrow$ 0.04}  \\
                                     & Ideal                   & 81.01 & \multicolumn{1}{c|}{}                      & 80.82 & \multicolumn{1}{c|}{}                      & 84.57 & \multicolumn{1}{c|}{}                      &   11.81 &            \\ \hline
\multirow{2}{*}{\begin{tabular}[c]{@{}l@{}}"-Sep"\\ ($w/$ ICL)\end{tabular}}  & Random                  & 85.14 & \multicolumn{1}{c|}{\multirow{2}{*}{\textbf{$\uparrow$ 2.86}}} & 82.82 & \multicolumn{1}{c|}{\multirow{2}{*}{\textbf{$\uparrow$ 1.53}}} & 88.63 & \multicolumn{1}{c|}{\multirow{2}{*}{\textbf{$\uparrow$ 0.26}}} &   63.08 & \multirow{2}{*}{\textbf{$\uparrow$ 14.03}} \\
                                     & Ideal                   & 88.00 & \multicolumn{1}{c|}{}                      & 84.35 & \multicolumn{1}{c|}{}                      & 88.89 & \multicolumn{1}{c|}{}                      &   77.11 &                        \\ \bottomrule
        \end{tabular}
        \label{tab:ideal_prompt}
    \end{subtable}

    \centering
    \begin{subtable}[t]{0.99\linewidth}
        \renewcommand\arraystretch{1.2}
        \caption{Ablations of In-Context Enhancing and In-Context Labeling on the "-Sep" and "-Cat" variants of PIC++.}
        \setlength\tabcolsep{1.75mm}
        \centering
        \begin{tabular}{l|cc|ccc}
            \toprule
             Variants                  & ICE          & ICL        & ShapeNetPart mIoU$\uparrow$ & Human3D mIoU$\uparrow$ & BEHAVE mIoU$\uparrow$ \\ \hline
             \multirow{4}{*}{"-Cat"} & -            & -          & 64.52        & 72.79   & 83.66  \\
                                       & \Checkmark   & -          & 75.15        & \textbf{82.03}   & 86.48  \\
                                       & -            & \Checkmark & 77.75        & 50.76   & 81.53  \\
                                       & \cellcolor{gray!10} \Checkmark   & \cellcolor{gray!10} \Checkmark & \cellcolor{gray!10} \textbf{84.13}        & \cellcolor{gray!10} 80.11   & \cellcolor{gray!10} \textbf{88.03}  \\ \hline
            \multirow{4}{*}{"-Sep"} & -            & -          & 79.22        & 74.74   & 84.03  \\
                                       & \Checkmark   & -          & 80.16        & 80.85   & 84.62  \\
                                       & -            & \Checkmark & 78.31        & 77.37   & 86.53  \\
                                       & \cellcolor{gray!10} \Checkmark   & \cellcolor{gray!10} \Checkmark & \cellcolor{gray!10} \textbf{85.14}        & \cellcolor{gray!10} \textbf{82.82}   & \cellcolor{gray!10} \textbf{88.63}  \\ \bottomrule
        \end{tabular}
        \label{tab:auxiliary_labeling}
    \end{subtable}
    
    \centering
    \begin{subtable}[t]{0.99\linewidth}
        \renewcommand\arraystretch{1.1}
        \caption{Different loss functions.}
        \setlength\tabcolsep{5mm}
        \centering
         \begin{tabular}{c|c|ccc}
                    \toprule
                    \multicolumn{1}{l|}{\#} & \multicolumn{1}{c|}{\begin{tabular}[c]{@{}c@{}}Loss\\ function\end{tabular}} & \begin{tabular}[c]{@{}c@{}}ShapeNetPart\\ mIoU$\uparrow$\end{tabular} & \begin{tabular}[c]{@{}c@{}}Human3D\\ mIoU$\uparrow$\end{tabular} & \begin{tabular}[c]{@{}c@{}}BEHAVE\\ mIoU$\uparrow$\end{tabular} \\ \hline
                    1     & $CD$    &  84.86    &  82.60    & 88.53        \\
                    2     & $Smooth-\ell_{1}$    & 84.44  & 81.88  & \textbf{88.64}        \\ 
                    \rowcolor{gray!10}
                    3     & $CD + Smooth-\ell_{1}$   &  \textbf{85.14}    & \textbf{82.82}    &  88.63  \\ \bottomrule
                \end{tabular}
                \label{tab:loss_function_2}
    \end{subtable}
    \label{tab:ablation_segmentation}
\end{table*}

\subsubsection{Ablations on PIC}
\noindent \textbf{Effectiveness of Joint Sampling.}
We investigate whether the JS module has a valid effect on four tasks. As shown in Tab.~\ref{tab:JSmodule}, both PIC-Sep and PIC-Cat face a sharp decline in performance on the four tasks when the JS module is absent, and they even fail to achieve the most basic goal of reconstructing masked tokens. These findings validate our intuition that maintaining the consistency of the input and target token sequence positions is an indispensable design. That is to say, the JS module is a simple yet effective way to compensate for the missing positional information.

\begin{table*}[h]
\centering
\setlength\tabcolsep{1.7mm}
\caption{\textbf{Ablation experiments with different label point bank sizes $N_{B}$.} S., H., B., and A. represent ShapeNetPart, Human3D, BEHAVE, and AKB-48, respectively. We report the ins. mIoU metric.}
\begin{tabular}{l|cccc|cccc|cccc}
\toprule
   & \multicolumn{4}{c|}{$N_{B}=8$}                    & \multicolumn{4}{c|}{$N_{B}=20$}                   & \multicolumn{4}{c}{$N_{B}=50$}                  \\ \hline
\multicolumn{1}{l|}{} & S. & H. & B. & A. & S. & H. & B. & A. & S. & H. & B. & A. \\
PIC++                & 85.14        & 82.82       & 88.63      & 63.08      & 84.37        & 81.28       & 88.42      & 64.47      & 82.73        & 70.52       & 85.89      & 61.74  \\
\bottomrule
\end{tabular}
\label{tab:nb_ablation}
\end{table*}

\noindent \textbf{Influence of Prompt Selecting Strategies.}
We explore the impact of the prompt selection on the model's prediction results. For the random selection method, during testing, we randomly select a pair of input-target point clouds from the training set that performs the same task as the query point cloud, serving as a task prompt. For the class-aware selection method, based on the previous one, we further select point clouds belonging to the same category as the query, such as airplanes, tables, chairs, etc. Additionally, we further delve into selecting two alternative prompts. To select pairs of examples that are paired with the query point cloud, we consider two factors: the Chamfer Distance~(CD)~\cite{chamferdistance} between the prompt and the query point cloud, and the feature similarity between them (features are extracted from pre-trained PointNet~\cite{pointnet}), which are respectively denoted as CD-aware and Fea-aware. As depicted in Tab.~\ref{tab:prompt_selecting}~(a)(b), the CD-aware method demonstrates the best performance and even outperforms the task-specific models on the registration task. Note that we report the random method in the main results~(Tab.~\ref{tab:main_result_1}), which means our model has a higher ceiling. This provides us a great opportunity to improve downstream task results by selecting higher-quality prompts, which will be the direction of future work.

\noindent \textbf{Effects of Mask Ratio.}
We conduct ablation experiments on the mask ratio at a wide range~(20$\%$-70$\%$). As shown in Tab.~\ref{tab:mask_ratio}, training our PIC with a lower mask ratio weakens its performance across various tasks, especially on the mask ratio of 20$\%$. Meanwhile, the best results in the four tasks are distributed at different mask ratios, but considering all downstream tasks as a whole, the model can achieve the highest performance when the mask ratio is 70$\%$. Unlike language data, we also find that keeping sparsity in training is necessary for mask point modeling for in-context learning. We find similar results as in MAE~\cite{mae}, and a higher mask ratio is required to ensure the model can learn hidden features well.

\noindent \textbf{Ablation on Point Sampling.}
We study how the sampling method used by PIC in JS modules affects the performance of the model. We use the two most common sampling methods: Farthest Point Sampling (FPS)~\cite{pointnet}, and Random Sampling (RS)~\cite{rs}. As shown in Tab.~\ref{tab:ablation_pic}~(a), for the tasks of reconstruction, denoising, and registration, FPS produces better results than RS. Especially for point cloud reconstruction, where FPS can collect more key points than RS, and these key points can describe the original outline of the entire point cloud. However, in the task of part segmentation, the results of RS exceed those of FPS.

\noindent \textbf{Prompt Engineering.}
We investigate the influence of the prompt layout and query on the experimental results. For PIC-Sep, we set up two layout options: one with the prompt before the query and the other with the prompt after the query. As shown in Tab.~\ref{tab:ablation_pic}~(b), the performance between the two designs has negligible differences. We simply choose the ``before" option to align with 2D in-context learning works.

\noindent \textbf{Loss Function.}
We explore to determine which loss function is most suitable for our PIC model. During training, we experiment with using Chamfer Distance $\ell_{1}$, $\ell_{2}$, and a combination of $\ell_{1}$ and $\ell_{2}$ as the loss functions for our PIC-Sep. As Tab.~\ref{tab:ablation_pic}~(c) shows, $\ell_{2}$ achieves the best result on all four tasks.

\subsubsection{Ablations on PIC++}

\noindent \textbf{Influence of In-Context Labeling.}
We conduct an in-depth investigation into the impact of In-Context Labeling (ICL) on our PIC model by comparing its performance with and without the ICL training strategy. We experiment with two prompt selection strategies: random prompts and ideal prompts. The ideal prompt refers to using the query point cloud and its corresponding target point cloud as its prompt, which is technically the ideal prompt and can lead to the best performance of the model. As shown in Tab.~\ref{tab:ablation_segmentation}~(a), by comparing the mIoU distances between using random and ideal prompts, we observe that the model trained with the ICL strategy has a higher upper limit of segmentation performance and greater potential.

\noindent \textbf{Effectiveness of In-Context Labeling and Enhancing.}
We also conduct detailed ablation experiments on ICL and ICE. As shown in Tab.~\ref{tab:ablation_segmentation}~(b), both ICL and ICE are essential for enhancing the segmentation performance of PIC++. 
We observe that using only the ICL training strategy increases the training challenge and may even lead to a decrease in segmentation performance due to introducing a significant amount of randomness. 
On the other hand, the ICE training strategy, which involves incorporating auxiliary training point cloud pairs, reinforces the model's learning of mapping relationships in prompts. 
Therefore, we argue that the best choice is to use ICE and ICL training strategies simultaneously.
This can maximize the segmentation performance of the model under multi-dataset joint training.

\noindent \textbf{Ablation on Different Sizes of Label Point Bank ($N_{B}$).}
We conducted ablation experiments with different $N_B$ values. As shown in Tab.~\ref{tab:nb_ablation}, although increasing the bank size to 20 and 50 causes a slight decline in performance, the model maintains competitive results; for instance, the score on ShapeNetPart only decreases marginally from 85.14 to 84.37 ($N_B=20$) and 82.73 ($N_B=50$), while AKB-48 even sees an improvement to 64.47 at $N_B=20$. These results demonstrate that our Dynamic In-Context Labeling strategy ensures that the model learns the underlying mapping rules rather than memorizing specific coordinates, allowing it to generalize effectively to new samples with more parts than those seen during training. Consequently, our method allows for the use of a larger $N_B$ in practical applications to guarantee that the number of part labels does not overflow, without compromising performance significantly.

\noindent \textbf{Loss Function.} We conduct ablation experiments on the loss function of PIC++. As shown in Tab.~\ref{tab:ablation_segmentation}~(c), compared with using CD loss or $Smooth-\ell_{1}$ loss alone, the best results can be obtained by combining the two.
In the part segmentation task, $Smooth-\ell_{1}$ is more appropriate for supervised learning of point-level prediction results.
Meanwhile, CD loss constrains the prediction of the auxiliary training task (ICE), allowing the model to learn more diverse knowledge about mapping relationships from the prompt pair.

\section{Conclusion}
\label{sec:conclusion}
We introduce Point-In-Context (PIC), a foundational framework designed to unify 3D point cloud understanding through in-context learning. To mitigate information leakage in masked point modeling, we propose a Joint Sampling module and establish a comprehensive multitask benchmark. PIC demonstrates robust performance across reconstruction, denoising, registration, and segmentation, showing strong capability in generalization.
Crucially, we extend this foundation to PIC++, which incorporates In-Context Labeling and In-Context Enhancing strategies to resolve the limitations of fixed label assignments. These contributions enable seamless scalability across diverse segmentation datasets and facilitate generalization to novel categories, achieving state-of-the-art performance on the proposed multi-entity benchmark. Collectively, PIC and PIC++ pave the way for generalist 3D vision models, marking a substantial step from task-specific tuning toward a scalable, in-context learning paradigm.

\noindent \textbf{Broad Impact.} Our work is the first to explore in-context learning for 3D point cloud understanding and is a very relevant but underexplored problem, including task definition, benchmark, and baseline models. 
Besides, we hope the setup of in-context learning in 3D and the curation of the in-context learning dataset is helpful to the community.

\section{Limitations}
\label{sec:limitations}
Although the proposed PIC and PIC++ demonstrate strong generalization and versatility, certain limitations remain. First, our experiments are conducted solely on object-level point clouds, leaving scene-level point cloud applications unexplored. Second, our framework is designed to process only 3D point clouds with XYZ coordinates, overlooking other modalities such as color and normal vectors, which restricts its applicability in multimodal scenarios. Finally, while contextual labels significantly enhance the model's generalization ability, they introduce randomness (random labels) during training, potentially leading to instability. Although our contextual augmentation strategy largely mitigates this issue, occasional randomness may still occur.

\section{Future Work}
\label{sec:futurework}
Future Work: First, we plan to extend our PIC framework to large-scale scene-level point clouds, aiming to develop a more comprehensive context learning framework for 3D point clouds. Second, we will incorporate additional modalities to enhance contextual learning and generalization, further exploring the application to broader tasks such as 3D detection and captioning.

\section*{Declarations}

\begin{itemize}
\item Funding: This work was supported by National Natural Science Foundation of China (No. 62473007), Guangdong Outstanding Youth Fund (No. 2026B1515020015), Shenzhen Innovation in Science and Technology Foundation for The Excellent Youth Scholars (No. RCYX20231211090248064).
This study is also supported under the RIE2020 Industry Alignment Fund Industry Collaboration Projects (IAF-ICP) Funding Initiative, as well as cash and in-kind contributions from the industry partner(s). 
It is also supported by Singapore MOE AcRF Tier 2 (MOE-T2EP20221-0011). It is also supported by the interdisciplinary doctoral grants (iDoc 2021-360) from the Personalized Health and Related Technologies (PHRT) of the ETH domain.
\item Data availability: The data used in the paper will be available at \url{https://github.com/fanglaosi/Point-In-Context}.
\item Code availability: The code used in the paper will be available at \url{https://github.com/fanglaosi/Point-In-Context}.


\end{itemize}


\bibliography{main}

\end{document}